\documentclass[10pt, jou, floatsintext]{apa7}
\usepackage{amssymb}
\usepackage{graphicx}
\usepackage[outdir=./]{epstopdf}
\usepackage{caption}
\usepackage{mathtools}
\usepackage{enumerate}
\usepackage{textcomp}
\usepackage{lingmacros}
\usepackage{apacite}
\usepackage{listings}
\usepackage{multirow}
\usepackage{stfloats}
\usepackage{todonotes}
\usepackage{svg}
\usepackage{booktabs}
\usepackage{stmaryrd}

\newcommand*{\bluesquare}[0]{%
\mathord{
        \includegraphics[
        height=.6em,
        width=.6em,
        keepaspectratio,
        ]{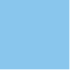}%
    }
}

\newcommand*{\orangecircle}[0]{%
\mathord{
        \includegraphics[
        height=.6em,
        width=.6em,
        keepaspectratio,
        ]{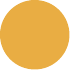}%
    }%
}

\usepackage{soul}

\newcommand{\den}[1]{\ensuremath{\llbracket #1 \rrbracket}}

\usepackage{lipsum}

\graphicspath{{./figures/}}
 
 \definecolor{Green}{RGB}{10,200,100}
  \definecolor{Red}{RGB}{200,100,50}

\makeatother

\title{From partners to populations: \\[.1em] A hierarchical Bayesian account of coordination and convention}
\shorttitle{Conventions}
\leftheader{Hawkins et al.}
\authorsnames[{*1}, {2}, {3}, {1}, {4}, {1,5}, {3,6}]{Robert D. Hawkins, Michael Franke, Michael C. Frank, \\Adele E. Goldberg, Kenny Smith, Thomas L. Griffiths, Noah D. Goodman}

\authorsaffiliations{{Department of Psychology, Princeton University}, 
{Institute for Cognitive Science, University of Osnabr\"uck},
{Department of Psychology, Stanford University},
{Centre for Language Evolution, University of Edinburgh},
{Department of Computer Science, Princeton University},
{Department of Computer Science, Stanford University }}

\abstract{Languages are powerful solutions to coordination problems: they provide stable, shared expectations about how the words we say correspond to the beliefs and intentions in our heads.
Yet language use in a variable and non-stationary social environment requires linguistic representations to be flexible: old words acquire new \emph{ad hoc} or partner-specific meanings on the fly. 
In this paper, we introduce CHAI (Continual Hierarchical Adaptation through Inference), a hierarchical Bayesian theory of coordination and convention formation that aims to reconcile the long-standing tension between these two basic observations.
We argue that the central computational problem of communication is not simply transmission, as in classical formulations, but continual \emph{learning} and \emph{adaptation} over multiple timescales.
Partner-specific common ground quickly emerges from social inferences within dyadic interactions, while community-wide social conventions are stable priors that have been abstracted away from interactions with multiple partners.
We present new empirical data alongside simulations showing how our model provides a computational foundation for several phenomena that have posed a challenge for previous accounts: (1) the convergence to more efficient referring expressions across repeated interaction with the same partner, (2) the gradual transfer of partner-specific common ground to strangers, and (3) the influence of communicative context on which conventions eventually form.
}


\authornote{This manuscript is based in part on non-archival work presented at the 39th, 40th, and 42nd Conferences of the Cognitive Science Society \cite{hawkins_convention-formation_2017,hawkins_emerging_abstractions_2018,hawkins2020generalizing}. Materials and code for reproducing all model simulations, behavioral experiments, and analyses are open and available online at \href{https://github.com/hawkrobe/conventions_model}{https://github.com/hawkrobe/conventions\_model}.\\
$^*$Correspondence should be addressed to Robert  Hawkins, e-mail: rdhawkins@princeton.edu}

\begin{document}
\maketitle

To communicate successfully, speakers and listeners must share a common system of semantic meaning in the language they are using. 
These meanings are \emph{social conventions} in the sense that they are arbitrary to some degree, but sustained by stable expectations that each person holds about others in their community \cite{lewis_convention:_1969,bicchieri_grammar_2006, hawkins2019emergence}.
Importantly, these expectations extend to complete strangers.
An English speaker may order an ``espresso'' at any café in the United States and expect to receive (roughly) the same kind of drink.
At the same time, meaning can be remarkably flexible and \emph{partner-specific}.
The same words may be interpreted differently by different listeners, or take on new \emph{ad hoc} senses over the course of a conversation \cite{clark_using_1996}. 
Interactions between friends and colleagues are filled with proper names, technical jargon, slang, shorthand, and inside jokes, many of which are unintelligible to outside observers.

The tension between these two basic observations, global stability and local flexibility, has posed a challenging and persistent puzzle for theories of convention.
Many influential computational accounts explaining how stable social conventions emerge in populations  do not allow for partner-specific meaning at all \cite<e.g.>{hurford1989biological,shoham1997emergence,barr_establishing_2004,skyrms2010signals,steels2011modeling,young_evolution_2015}.
These accounts typically examine groups of interacting agents who update their representations of language after each interaction.
While the specific update rules range from simple associative mechanisms \cite<e.g.>{steels_self-organizing_1995} or heuristics \cite<e.g>{Young96_EconomicsOfConvention} to more sophisticated deep reinforcement learning algorithms \cite<e.g.>{tieleman2019shaping,graesser2019emergent,mordatch2017emergence}, all of these accounts assume that agents update a single, monolithic representation of language to be used with every partner, and that agents do not (knowingly) interact repeatedly with the same partner.

Conversely, accounts emphasizing rapid alignment \cite{pickering2004toward} or the development of partner-specific common ground \cite{ClarkMarshall1981,ClarkWilkesGibbs86_ReferringCollaborative} across extended interactions with the same partner typically do not specify mechanisms by which community-wide conventions arise over longer timescales.
The philosopher Donald Davidson articulated one of the most radical of these accounts.
According to \citeA{davidson1984communication,davidson_nice_1986,davidson1994social}, while we bring background expectations (``prior theories'') into interactions, it is the ability to coordinate on \emph{partner-specific} meanings (``passing theories'') that is ultimately responsible for communicative success: 
\begin{quote}
\footnotesize\emph{In order to judge how he will be interpreted, [the speaker] uses a picture of the interpreter’s readiness to interpret along certain lines, [...] the starting theory of interpretation. 
As speaker and interpreter talk, their ``prior'' theories become more alike; so do their ``passing'' theories. 
The asymptote of agreement and understanding is when passing theories coincide. 
Not only does it have its changing list of proper names and gerrymandered vocabulary, but it includes every successful use of any other word or phrase, no matter how far out of the ordinary [...] 
Such meanings, transient though they may be, are literal. \\\cite[p.~261]{davidson_nice_1986}.}
\end{quote}
This line of argument led \citeA{davidson_nice_1986} to conclude that ``there is no such thing as a language'' (p.~265), and to abandon appeals to convention altogether (see \citeNP{heck_idiolect,lepore2007reality,hacking1986nice,dummett1994} for further discussion of Davidson's view; \citeNP{armstrong2016coordination,armstrong2016problem}, provides a philosophical foundation for our synthesis).

In this paper, we propose an account of coordination and convention that aims to reconcile the emergence of community-level conventions with partner-specific common ground in a unified cognitive model.
This theory is motivated by the computational problems facing individual agents who must communicate with one another in a variable and non-stationary world. 
We suggest that three core cognitive capacities are needed for an agent to solve this problem:
\begin{description}
\item[C1:] the ability to represent \textbf{variability} about what words will mean to different partners,
\item[C2:] the ability to coordinate on partner-specific meanings via flexible \textbf{online learning}, and
\item[C3:] the ability to gradually \emph{generalize} stable expectations about meaning from individual interactions.
\end{description}
These properties are naturally formalized in a hierarchical Bayesian framework, which we call CHAI (Continual Hierarchical Adaptation through Inference). 
Indeed, one of our central theoretical aims is to ground the problem of convention formation --- a fundamentally interactive, social phenomenon --- in the same domain-general cognitive mechanisms supporting learning in other domains where abstract, shared properties need to be inferred along with idiosyncratic particulars of instances \cite{berniker2008estimating,GoodmanUllmanTenenbaum11_TheoryOfCausality,tenenbaum_how_2011,kleinschmidt2015robust}.

Our argument is structured around a series of three key phenomena in the empirical literature that have proved evasive for previous theoretical accounts of coordination and convention: 
\begin{description}
\item[P1:] the convergence to \textbf{increasingly efficient referring expressions} over repeated interactions with a single partner,
\item[P2:] the transition from \textbf{partner-specific pacts} to communal conventions that are expected to generalize to new partners, and
\item[P3:] the influence of \textbf{communicative context} on which terms eventually become conventionalized 
\end{description}

We begin by introducing the \emph{repeated reference game} paradigm at the center of this literature and reviewing the empirical evidence supporting each of these phenomena.
We then introduce CHAI in detail and highlight several important properties of our formulation.
The remainder of the paper proceeds through each of the three phenomena (\textbf{P1}-\textbf{P3}) in turn. 
For each phenomenon, we present computational simulations to evaluate how CHAI explains existing data, and introduce data from new real-time, multi-player behavioral experiments to test novel predictions when existing data does not suffice.
Finally, we close by discussing several broader consequences of the theory, including the continuity of language acquisition and convention formation in adulthood and domain-generality of discourse processes, as well as several limitations, addressing questions of scalability and incrementality.

\subsection{Three lessons about convention formation from repeated reference games}

A core function of language is \emph{reference}: using words to convey the identity of an entity or concept. 
Loosely inspired by \citeA{wittgenstein2009philosophical}, empirical studies of coordination and convention in communication have predominantly focused on the subset of language use captured by simple ``reference games.''
In a reference game, participants are assigned to speaker and listener roles and shown a context of possible referential targets (e.g. images).
On each trial, the speaker is asked to produce a referring expression --- typically a noun phrase --- that will allow the listener to select the intended target object from among the other objects in the context.

Critically, unlike typical studies of referring expression generation \cite{van_deemter_computational_2016,degen2020redundancy,dale1995computational}, \emph{repeated reference games} ask speakers to refer to the same targets multiple times as they build up a shared history of interaction with their partners (see Table \ref{table:parameters} in Appendix for a review of different axes along which the design has varied). 
And unlike agent-based simulations of convention formation on large networks \cite<e.g.>{steels2011modeling,barr_establishing_2004,centola_spontaneous_2015}, which typically match agents with a new, anonymous partner for each trial, repeated reference games ensure that participants know their partner's identity and maintain the same partner throughout extended interactions.
This design allows us to observe how the speaker's referring expressions for the same objects change as a function of interaction with that particular partner.
We now highlight three findings of particular theoretical significance that emerge from the repeated reference paradigm. 

\paragraph{P1:~Increasingly efficient conventions}
The most well-known phenomenon observed in repeated reference games is a dramatic reduction in message length over multiple rounds \cite{krauss_changes_1964,ClarkWilkesGibbs86_ReferringCollaborative, hawkins2020characterizing}. 
The first time participants refer to a figure, they tend to use a lengthy, detailed description (e.g. ``the upside-down martini glass in a wire stand'') but with a small number of repetitions --- between 3 and 6, depending on the pair of participants --- the description may be cut down to the limit of just one or two words (``martini'')\footnote{Of course, referring expressions are also lengthened for many reasons other than pure reference, such as politeness (\textit{Professor Davidson vs. Don}), affection (\textit{Donny vs Don}), emphasis (\textit{the one and only}), or any number of manner implicatures \cite<see>{horn1984toward,levinson2000presumptive}. However, the marked meanings of these longer forms are only obtained against the backdrop of an unmarked or ``default'' form; repeated reference games set these other functions aside to examine where unmarked expectations come from and how they depend on discourse context \cite{grosz1974structure,grosz1986attention}. This distinction may be seen by considering the non-referential implicatures that may be triggered if a speaker suddenly switched from ``martini'' back to their original longer description at the end of a game.}.
These final messages are as short or shorter than the messages participants produce when they are instructed to generate descriptions for themselves to interpret in the future \cite{FussellKrauss89_IntendedAudienceCommonGround} and are often incomprehensible to overhearers who were not present for the initial messages \cite{SchoberClark89_Overhearers}.
This observation sets up a first puzzle of \emph{ad hoc} convention formation in dyads:
How does a word or short description that would be largely ineffective at the outset of a conversation take on local meaning over mere minutes of interaction?
\begin{figure}[t!]
\centering
\includegraphics[scale=1.3]{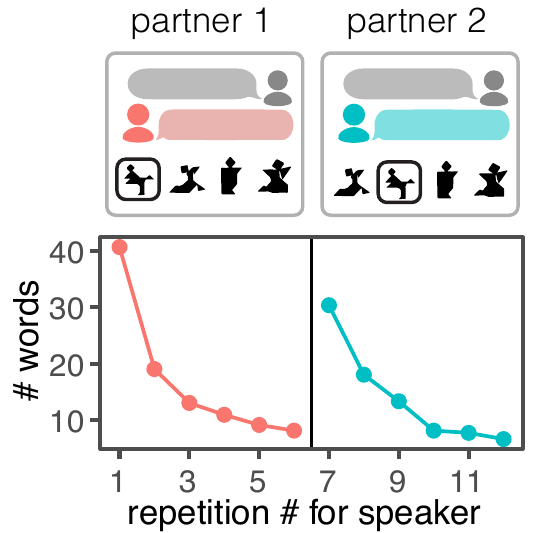}
\vspace{1em}
\caption{\textit{Classic phenomena in repeated reference games.} Over multiple iterations with the same partner, speakers speakers converge on increasingly efficient referring expressions (reps.~1-6). When the listener is replaced by a new, naive partner, speakers display a key signature of partner-specificity, reverting to longer utterances before converging again with their new partner (reps.~7-12). Comprehension failures tend to be rare ($\sim 2.3\%$) throughout the experiment, indicating that speakers modulate their utterances effectively. Data from Table 3 in \protect\citeA{wilkes-gibbs_coordinating_1992}.}
\label{fig:clark92}
\end{figure}

\paragraph{P2:~Partner-specific conventions}
Because meaning is grounded in the evolving common ground shared with each partner, \emph{ad hoc} conventions established over a history of interaction with one partner are not necessarily transferred to other partners \cite{metzing_when_2003,weber_cultural_2003,brown2009partner}\footnote{We use the term ``\emph{ad hoc} convention'' \cite<inspired by>{Barsalou83_AdHocCategories} interchangeably with the more common ``conceptual pact''   \cite{BrennanClark96_ConceptualPactsConversation,IbarraTanenhaus16_FlexibilityConceptualPacts} to emphasize the theoretical relationship between this construct and the usual sense of convention referring to longer-term communal knowledge.}. 
For example, \citeA{wilkes-gibbs_coordinating_1992} paired participants for a standard repeated reference game, but after six rounds, the listener was replaced by a naive partner. 
Without partner-specific representations, we would expect speakers to continue using the short labels they had converged on with their first partner; instead, speakers reverted to the longer utterances they had initially used, and then coordinated on new \emph{ad hoc} conventions with their new partner (see Fig.~\ref{fig:clark92}).
These effects raise our second puzzle: how do community-level conventions form in the presence of such strong partner-specificity?
When are agents justified in transferring an \emph{ad hoc} convention formed with one partner to a new, unseen partner? 

One important empirical clue was provided by \citeA{fay_interactive_2010}, who examined the emergence of conventions in a lab experiment where communities of eight people played a repeated  graphical communication game similar to Pictionary, where participants produced drawings to allow their partner to identify a concept from a list of possibilities. 
The 8 participants in each network interacted dyadically with every other member of the community, in turn, for a series of seven repeated reference games. 
Strikingly, participants behaved as observed by \citeA{wilkes-gibbs_coordinating_1992} during the first few partner swaps, consistent with partner-specificity, but with subsequent partners, their initial drawings showed a gradual convergence with the  conventionalized drawings they had settled upon with previous partners, indicating a slow gradient of generalization within their community.

While intriguing, this work was limited by an extremely small sample size~($N=4$~groups)~and technical challenges facing the measurement of conventions in the graphical modality \cite<see also>{hawkins2019disentangling}.
More recent work has adopted a similar design for an artificial-language communication task \cite{raviv2019larger} but collapses across repeated dyadic interactions to exclusively analyze network-level metrics, making it difficult to assess partner-specificity.
Given these limitations of existing data, we evaluate our model's predictions using new data from a large-scale, real-time web experiment directly extending \citeA{wilkes-gibbs_coordinating_1992} to larger networks.

\paragraph{P3:~Context-sensitive conventions}

Finally, while a degree of arbitrariness is central to conventionality -- there must exist more than one solution that would work equally well -- this does not necessarily imply that all possible conventions for a given meaning are equally likely in practice, or even any convention will form at all \cite{HawkinsGoldstone16_SocialConventions}.
Indeed, functional accounts of language have frequently observed that lexical systems are well-calibrated to the needs of users under the statistics of their communicative environment \cite{gibson2019efficiency}.
This Optimal Semantic Expressivity  hypothesis \cite<OSE;>{frankblogpost} has held remarkably well for the lexical distributions found in natural languages across semantic domains like color words and kinship categories \cite{KempRegier12_KinshipCategories,regier201511,gibson2017color,kemp2018semantic}.

While such long-term, diachronic sensitivity to context has been explained by abstract principles of optimality, such as the equilibria concepts of evolutionary game theory \cite{jager2007evolution,jager2007language}, it has not yet been grounded in a cognitive and mechanistic account of the immediate, synchronic processes unfolding in the minds of individual agents while they interact.
In other words, while there is abundant empirical evidence for context-sensitivity in the \emph{outcomes} of convention formation processes, our third puzzle concerns which cognitive mechanisms  in individuals may be necessary or sufficient to give rise to such conventions \cite<see>[which raises a similar linking problem]{brochhagen2021brief}.

Repeated reference games have emerged as a promising method for probing these mechanisms in the lab. 
Such games allow researchers to  explicitly control the communicative context and observe the resulting distribution of conventions that emerge when participants communicate using artificial languages \cite{WintersKirbySmith14_LanguagesAdapt, KirbyTamarizCornishSmith15_CompressionCommunication,winters2018contextual} or natural language \cite{hawkins2020characterizing}.
While these studies are informative, it has remained challenging to directly evaluate cognitive models against the \emph{full trajectories} of convention formation on a trial-by-trial basis.
In our final section, we report new empirical data from a dyadic repeated reference task manipulating context, where simulated agents and human participants are shown directly analogous trial sequences.

\section{Convention formation as Hierarchical Bayesian inference}

In this section, we propose a unified computational account of \emph{ad hoc} coordination and convention formation that aims to address these three empirical puzzles. 
We begin from first principles: What is the core computational problem that must be solved to achieve successful communication?
Classically, this problem has been formulated in terms of coding and compression \cite{Shannon48}. 
An intended meaning in the speaker's mind must be encoded as a signal that is recoverable by the receiver after passing through a noisy transmission channel.
This transmission problem has since been enriched to account for \emph{pragmatics} -- the ability of speakers and listeners to use context and social knowledge to go beyond the literal meaning of messages \cite{RosenbergCohen66_ReferentialProcesses,sperber1986relevance}.
We take the Rational Speech Act framework \cite<RSA;>{FrankGoodman12_PragmaticReasoningLanguageGames,goodman_pragmatic_2016,FrankeJager16_ProbabilisticPragmatics} as representative of this current synthesis, formalizing communication as recursive social inference in a probabilistic model (see Appendix A for technical details.)
In the next section, we review this basic framework and raise two fundamental computational problems facing it.
These problems motivate the introduction of continual learning in the CHAI model.

\subsection{Models of communication with static meaning}

For concreteness, we restrict our scope to reference in a context $\mathcal{C}$ containing a discrete set of objects $o\in\mathcal{O}$, but the same formulation aims to apply more generally.
In this referential setting, the RSA framework defines a pragmatic speaker, denoted by $S_1$, who must choose an utterance $u$ that will allow their partner to choose a particular target object $o^* \in \mathcal{C}$.
They attempt to satisfy Gricean Maxims \cite{Grice75_LogicConversation} by selecting utterances according to a utility function $U(u;o)$ that balances informativity to an imagined listener against the cost of producing an utterance.
Specifically, $S_1$ chooses from a ``softmax distribution'' concentrating mass on the utterance that maximizes $U(u;o)$ to an extent modulated by a free parameter $\alpha_S \in[0,\infty]$:
\begin{align}
S_1(u | o) & \propto   \exp\{\alpha_S \cdot U(u; o)\}
\end{align}
For $\alpha_S = 1$, this decision rule corresponds to Luce's choice axiom \cite{luce1959individual}. 
Larger settings of $\alpha_S$ concentrate more probability on the single utterance maximizing utility.

The basic speaker utility function in the RSA framework is defined as follows:
\begin{align}
U(u; o) & = (1-w_C) \cdot \underbrace{\log L_0(o | u)}_{\mathclap{\text{informativity}}} -\, w_C \cdot \underbrace{c(u)}_{\mathclap{\text{cost}}} \label{eq:RSAspeaker}
\end{align}
where $c(u)$ is a function giving the cost of producing $u$, assuming longer utterances are more costly, and $w_C \in [0,1]$ is a second free parameter controlling the relative weight of informativity and parsimony in the speaker's production.
Critically, the informativity term in Eq.~\ref{eq:RSAspeaker} is defined by how well $u$ transmits the intended target $o^*$ to an imagined listener.
The simplest imagined listener, $L_0$, is typically called the ``literal listener'' because they are assumed to identify the target relying only on the literal meaning of the received utterance, without appealing to further social reasoning about the speaker.
That is, the probability of the imagined listener choosing object $o$ is simply assumed to be proportional to the meaning of $u$ under some (static) lexical function $\mathcal{L}$:
\begin{align}
L_0(o | u) &\propto  \mathcal{L}(u,o)\nonumber
\end{align}
Throughout this paper, we will take $\mathcal{L}$ to be a traditional Boolean function evaluating whether or not the expression $u$ applies to the entity in question\footnote{Due to the current limitations of representing lexical meaning in formal semantics, it has not been straightforward to specify a truth-conditional function explaining listener behavior for natural-language utterances (e.g. what makes one drawing belong in the literal extension of ``upside-down martini glass'' but not another, when neither of them are literally martini glasses?) This representation is convenient for our simulations, where we consider all possible discrete mappings between utterances and objects in the context, but better representations of lexical meaning may be substituted \cite<see>{potts2019case}. For example, Appendix B works out an example using a real-valued, continuous function \cite{degen2020redundancy} such as those learned by multi-modal neural networks \cite{monroe_colors_2017,achlioptas2019shapeglot,hawkins2019continual}.}:
$$
\mathcal{L}(u,o) = \left\{ \begin{array} {rl} 1 & \textrm{if $o \in \den{u}$} \\ 0 & \textrm{otherwise} \end{array}\right.
$$


\subsection{Two fundamental problems for static meaning}

The RSA framework and its extensions provide an account for a variety of important phenomena in pragmatic language use \cite<e.g.>{Scontras_problang,KaoWuBergenGoodman14_NonliteralNumberWords,TesslerGoodman16_Generics,LassiterGoodman15_AdjectivalVagueness}.
Yet it retains a key assumption from classical models: that the speaker and listener must share the same literal ``protocol'' $\mathcal{L}$ for encoding and decoding messages.
In this section, we highlight two under-appreciated challenges of communication that complicate this assumption. 

The first problem arises from the existence of \emph{variability} throughout a language community \cite{kidd2018individual,wangidiosyncratic}. 
Different listeners may recover systematically different meanings from the same message, and different speakers may express the same message in different ways.
For example, doctors may fluently communicate with one another about medical conditions using specialized terminology that is meaningless to patients. 
The words may not be in the patient's lexicon, or common words may be used in non-standard ways.
That is, being fluent speakers of the same language does not ensure agreement for the relevant meanings expressed in every context. 
Different partners may be using different functions $\mathcal{L}$.

The second problem arises from the \emph{non-stationarity} of the world. 
Agents are continually presented with new thoughts, feelings, and entities, which they may not already have efficient conventions to talk about \cite{gerrig1988beyond}.
For example, when new technology is developed, the community of developers and early adopters must find ways of referring to the new concepts they are working on (e.g. \emph{tweeting}, \emph{the cloud}). 
Or, when researchers design a new experiment with multiple conditions, they must find ways of talking about their own \emph{ad hoc} abstractions, often converging on idiosyncratic names that can be used seamlessly in meetings.
That is, any fixed $\mathcal{L}$ shared by a group of speakers at one moment in time can quickly become outdated \cite<see>[for a demonstration of the related problems posed by non-stationary for large neural language models]{lazaridou2021pitfalls}.
We must have some ability to extend our language on the fly as needed.

\subsection{CHAI: A model of dynamic meaning}

Rather than assuming a monolithic, universally shared language, we argue that agents solve the fundamental problems posed by variability and non-stationarity by attempting to continually, adaptively \emph{infer} the system of meaning used by their current partner.
When all agents are continually learning in this way, and changing their own behavior to best respond, we will show that they are not only able to coordinate on local, \emph{ad hoc} meanings or pacts with specific partners but also abstract away \emph{conventions} that are expected to be shared across an entire community.
We introduce the CHAI (Continual Hierarchical Adaptation through Inference) model in three steps, corresponding to how it formalizes the three core capacities \textbf{C1-C3}: hierarchical uncertainty about meaning, online partner-specific learning, and inductive generalization.

\paragraph{C1:~Representing variability in meaning via structured uncertainty} 

When an agent encounters a communication partner, they must call upon some representation about what they expect different signals will mean to that partner. 
We therefore replace the monolithic, static function $\mathcal{L}$ with a \emph{parameterized family} of lexical meaning functions by $\mathcal{L}_{\phi}$, where different values of $\phi$ yield different possible systems of meaning. 
To expose the dependence on a fixed system of meaning, Eq.~\ref{eq:RSAspeaker} can be re-written to give behavior under a fixed value of $\phi$:
\begin{align}
L_0(o | u, \phi) &\propto  \mathcal{L}_\phi(u,o)\hfill\label{eq:RSA} \\
U(u; o, \phi) & = (1-w_C) \cdot \log L_0(o | u, \phi) -\, w_C \cdot c(u) \nonumber  \\
S_1(u | o,\phi) & \propto   \exp\{\alpha_S \cdot U(u; o, \phi)\} \nonumber 
\end{align}

While we will remain agnostic for now to the exact functional form of $\mathcal{L}_\phi$ and the exact parameter space of $\phi$, there are two computational desiderata we emphasize.
First, given the challenge of variability raised in the previous section, these expectations ought to be \emph{sensitive to the overall statistics of the population}. 
That is, an agent should know that more people will share the meaning of some words than others, and should conversely expect more consensus about how to refer to some concepts than others.
Second, expectations about which meanings will be evoked for a given utterance and which utterances are expected to be used to express a meaning should be \emph{sensitive to the social identity of one's partner}.

The first desideratum -- the ability to represent variability in the population -- motivates a \emph{probabilistic} formulation.
Instead of holding a single static function $\mathcal{L}_{\phi}$, which an agent assumes is shared perfectly in common ground (i.e. one $\phi$ for the whole population), we assume each agent maintains uncertainty over the exact meaning of each word as used by different partners.
In a Bayesian framework, this uncertainty is specified by a prior probability distribution $P(\phi)$ over possible function parameters.
For example, imagine a doctor giving a diagnosis to a new patient.
Under some possible values of $\phi$, a piece of medical jargon like ``sclerotic aorta'' refers unambiguously to the patient's heart condition.
Under other values of $\phi$, it has a less clear meaning. 
A doctor with good bedside manner should assign some probability to each possibility rather than assuming everyone will share the same precise meaning they learned in medical school. 
Importantly, this variability will be different for different words: likely more people share the meaning of ``dog'' than ``sclerotic aorta''.
This core idea of introducing uncertainty over a partner's lexical semantics has previously been explored in the context of one-shot pragmatic reasoning, where it was termed \emph{lexical uncertainty} \cite{BergenGoodmanLevy12_Alternatives,PottsLevy15_Or,bergen_pragmatic_2016,potts2016embedded}, as well as in the context of iterated dyadic interactions \cite{SmithGoodmanFrank13_RecursivePragmaticReasoningNIPS}. 

\begin{figure}[b!]
\centering
\includegraphics[scale=0.35]{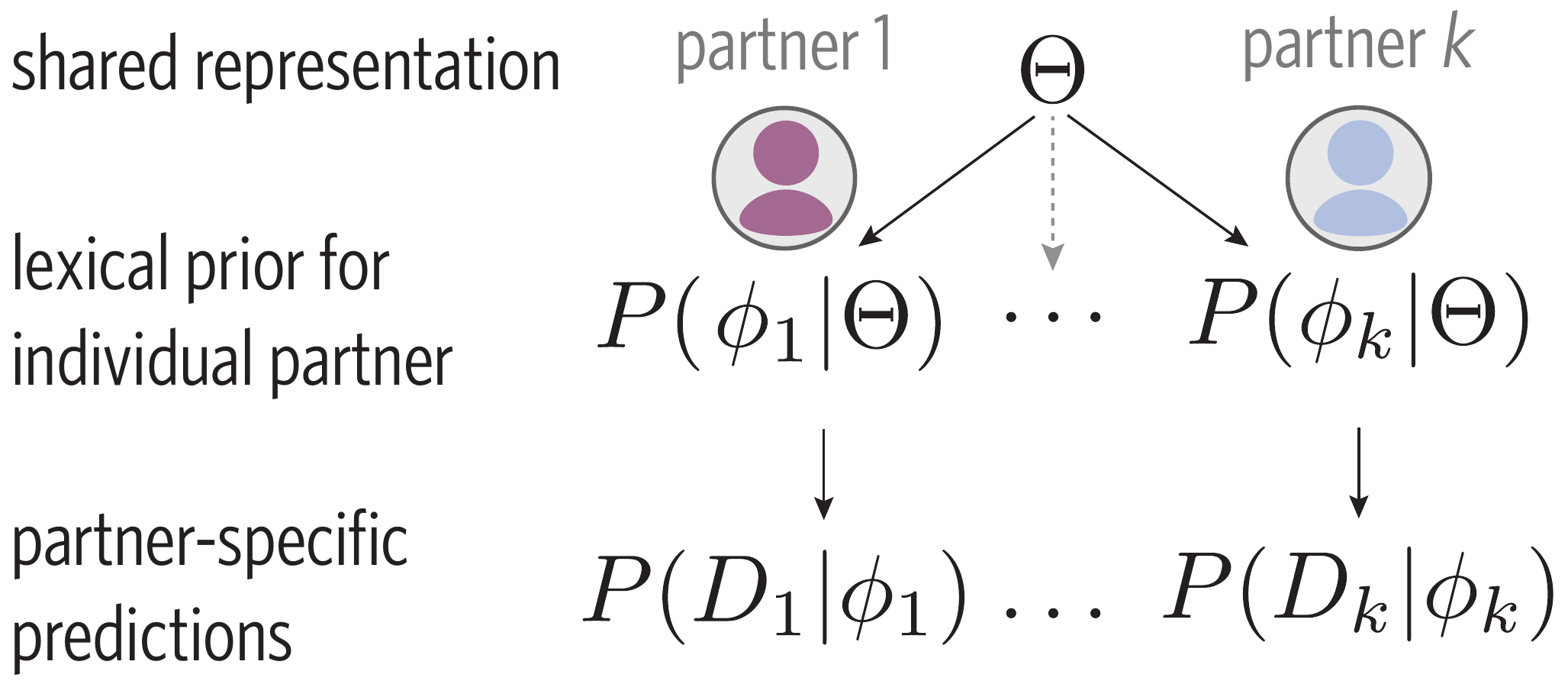}
\vspace{.5em}
\caption{\emph{Schematic of hierarchical model.} At the highest level, denoted by $\Theta$, is a representation of aspects of meanings expected to be shared across all partners. These conventions serve as a prior for the systems of meanings used by specific partners, $\phi_k$. Partner-specific representations give rise in turn to predictions about language use $P(D_k|\phi_k)$, where $D_k$ represents observations in a communicative interaction with partner $k$. By inverting this modeld, agents can adapt to \emph{local, ad hoc} conventions and gradually update their beliefs about conventions in their broader community.}
\label{fig:model_schematic}
\end{figure}

Second, this representation should also, in principle, be sensitive to the social identity of the partner: a doctor should be able to form different expectations about a new colleague than a new patient \cite{clark_communal_1998}.
This desideratum -- sensitivity to partner-specific meanings -- motivates a \emph{hierarchical} model, where uncertainty is represented by a multi-level prior. 
At the highest level of the hierarchy is \emph{community-level} uncertainty $P(\Theta)$, where $\Theta$ represents an abstract ``overhypothesis'' about the overall distribution of all possible partners. 
This level can be viewed as a representation of long-term ``communal lexicons'' about common knowledge based on community membership \cite{ClarkMarshall1981}. 
$\Theta$ then parameterizes the agent's \emph{partner-specific} uncertainty $P(\phi_{k} | \Theta)$, where $\phi_k$ represents the specific system of meaning used by partner $k$ (see Fig.~\ref{fig:model_schematic}). 
$\phi_k$ can be viewed as the ``idiolect'' that has been fine-tuned to account for partner-specific common ground and conceptual pacts from previous interactions\footnote{We focus for simplicity on this basic two-layer hierarchy, but the model can be straightforwardly extended to representing uncertainty at intermediate layers of social structure, including whether partners belong to distinct sub-communities \cite<e.g. represented by discrete latent variables>{GershmanEtAl17_StructureSocialInfluence,gershman2020social}, which may explain code-switching \cite{auer_code-switching_2013,hawkinsrespect} and other social inferences based on language use \cite{kinzler2021language,IsaacsClark87_ReferencesExpertsNovices,roberts_experimental_2010}.}.

To integrate lexical uncertainty into our speaker and listener models, we assume they each act in a way that is expected to be successful \emph{on average}, under likely values of $\phi_k$ \cite{SmithGoodmanFrank13_RecursivePragmaticReasoningNIPS}.
In other words, they sample actions by marginalizing over their own beliefs $P_S(\phi_k)$ or $P_L(\phi_k)$ about different meanings their partner $k$ may be using.
\begin{align}
L(o|u) &\propto   \exp\left\{\alpha_L\cdot \textstyle{\int} P_L(\phi_k)  \log S_1(u|o, \phi_k)\,\,d\phi_k\right\}\nonumber\\
S(u|o) &\propto  \exp\left\{\alpha_S \cdot \textstyle{\int} P_S(\phi_k)  U(u; o, \phi_k) \,\,d\phi_k\right\}\label{eq:marginalized}
\end{align}
where $\alpha_S, \alpha_L \in[0,\infty]$ control the speaker's and listener's soft-max optimality, respectively\footnote{We denote $L$ and $S$ without a subscript because they are the only speaker and listener models we use in simulations throughout the paper -- the subscripted definitions are internal constructs used to define these models -- but in the terminology of the RSA framework they represent $L_1$- and $S_1$-level pragmatic agents with lexical uncertainty. We found that higher levels of recursion were not  necessary to derive the phenomena of interest, but $L_n$ and $S_n$-level lexical uncertainty models may be generalized by replacing $S_1$ in the listener equation, and $L_0$ in the speaker's utility definition, with standard RSA definitions of $n-1$-level agents \cite<see also>{zaslavsky2020rate}.}.

\paragraph{C2: Online learning via partner-specific inference}

The formulation in Eq.~\ref{eq:marginalized} derives how agents ought to act under uncertainty about the lexicon being used by their partner, $P(\phi_k)$.
But how do beliefs about their partner change over time?
Although an agent may begin with significant uncertainty about the system of meaning their partner is using in the current context, further interactions provide useful information for reducing that uncertainty and therefore improving the success of communication.
In other words, \emph{ad hoc} convention formation may be re-cast as an inference problem.
Given observations $D_k$ from interactions with partner $k$, an agent can update their beliefs about their partner's latent system of meaning following Bayes rule:
\begin{equation}
\begin{array}{rcl}
\label{eq:joint_inference}
P(\phi_k, \Theta | D_k)  & \propto &  P(D_k | \phi_k, \Theta) P(\phi_k, \Theta) \\
                           & =   & P(D_k | \phi_k) P(\phi_k | \Theta) P(\Theta)
\end{array}
\end{equation}
This joint inference decomposes the partner-specific learning problem into two terms, a prior term $P(\phi_k | \Theta)P(\Theta)$ and a likelihood term $P(D_k | \phi_k)$.
The prior term captures the idea that, in the absence of strong evidence of partner-specific language use, the agent ought to regularize toward their background knowledge of conventions: the aspects of meaning that all partners are expected to share in common.
The likelihood term represents predictions about how a partner would use language in context under different underlying systems of meaning.

Importantly, the posterior obtained in Eq.~\ref{eq:joint_inference} allows agents to explicitly maintain \emph{partner-specific expectations}, as used in Eq.~\ref{eq:marginalized}, by marginalizing over community-level uncertainty:
\begin{equation}
P(\phi_k | D_k) =  \int_{\Theta}P(\phi_k, \Theta | D_k)  d\Theta
\end{equation}
We will show that when agents learn about their partner in this way, and adjust their own production or comprehension accordingly (i.e.~Eq.~\ref{eq:marginalized}), they are able to coordinate on stable \emph{ad hoc} conventions.

\paragraph{C3: Generalization to new partners via hierarchical induction}

The posterior in Eq.~\ref{eq:joint_inference} also provides an inductive pathway for partner-specific data to inform beliefs about community-wide conventions.
Agents update their beliefs about $\Theta$, using data accumulated from different partners, by marginalizing over beliefs about specific partners:
\begin{equation}
\begin{split}
    P(\Theta | D)  = & \int_{\phi} P(\phi, \Theta | D) d\phi \\
\end{split}
\end{equation}
where $D = \bigcup_{k=1}^N D_k$, $\phi = \phi_1 \times \dots \times \phi_N$, and $N$ is the number of partners previously encountered. 
Intuitively, when multiple partners are inferred to use similar systems of meaning, beliefs about $\Theta$ shift to represent this abstracted knowledge: it becomes more likely that novel partners in one's community will share it as well.
Note that this population-level posterior over $\Theta$ not only represents what the agent has learned about the central tendency of the group's conventions, but also the \emph{spread} or variability, capturing the notion that some word meanings may be more widespread than others.

The updated $\Theta$ should be used to guide the prior expectations an agent brings into a subsequent interactions with strangers.
This transfer is sometimes referred to as ``sharing of strength'' or ``partial pooling'' because pooled data is smoothly integrated with domain-specific knowledge.
This property has been key to explaining how the human mind solves a range of other difficult inductive problems in the domains of concept learning \cite{KempPerforsTenenbaum07_HBM, tenenbaum_how_2011}, causal learning \cite{KempPerforsTenenbaum07_HBM,KempGoodmanTenenbaum10_LearningToLearn},  motor control \cite{berniker2008estimating}, and speech perception \cite{kleinschmidt2015robust}.
One consequence is the ``blessing of abstraction,'' \cite{GoodmanUllmanTenenbaum11_TheoryOfCausality} where it is possible under certain conditions for beliefs about the community's conventions \emph{in general} to outpace beliefs about the idiosyncracies of individual partners \cite{gershman2017blessing}.

\subsection{Further challenges for convention formation}

The formulation in the previous section presents the core of CHAI.
Here, we highlight several additional features addressing more specific challenges raised by prior work on communication and which we will encounter in the simulations reported in the remainder of the paper. 
Our organization of these details is motivated by the analysis of \citeA{spike_minimal_2017}, who highlighted three common issues that all accounts of convention formation must address: (1) the form of feedback available, (2) the influence of memory and temporal discounting, and (3) the form of pragmatic reasoning being used.
Finally, we set up the basic simulation framework that will be used throughout the rest of the paper.

\paragraph{The role of social observation}

Learning and adaptation depend critically on the availability and quality of social observations $D_k$ (Eq.~\ref{eq:joint_inference}).
If the speaker has no way of probing the listener's understanding, or if the listener has no way of comparing their interpretation against the speaker's intentions, however indirectly, they can only continue to rely on their prior expectations, with no ground for conventions to form  \cite{HupetChantraine92_CollaborationOrRepitition,GarrodFayLeeOberlanderMacLeod07_GraphicalSymbolSystems}.
Communication is empirically hindered under degraded observation conditions  \cite{KraussWeinheimer66_Tangrams,KraussBricker67_Delay,KraussEtAl77_AudioVisualBackChannel,SchoberClark89_Overhearers}, and we have all been in situations where we thought we were on the same page with a partner and only realized that we misunderstood much later, when the consequences because clear.
In principle, we expect that $D_k$ should reflect all relevant sources of information that may expose an agent's state of understanding or misunderstanding.
Not just ostensive signals like pointing \cite{vandeBraak2021}, but verbal and non-verbal backchannels (e.g. \emph{mmhmm}, nods or quizzical looks), forms of self-initiated and other-initiated repair \cite<e.g. clarification questions or requests for confirmation>{SchegloffEtAl77_Repair, DingemanseEtAl15_RepairUniversal,arkel2020simple}, and downstream actions taken in the world (e.g. attempts to follow instructions).

While incorporating these richer sources of information presents an exciting line of future work, we restrict our scope to the feedback traditionally provided by the empirical repeated reference task, where the speaker's intended target and the listener's response are revealed at the end of each trial. 
Formally, this information can be written as a set of tuples $D_k = \{o^*, u', o'\}_{t=1}^T$, where $o^*$ denotes the speaker's intended target, $u'$ denotes the utterance they produced, and $o'$ denotes the listener's response, on each previous trial $t$.
To specify the likelihoods in Eq.~\ref{eq:joint_inference} for this referential setting, we assume each agent should infer their partner's lexicon $\phi_k$ by conditioning on their \emph{partner's} previous behavior.
The listener on a given trial should use the probability that a speaker would produce $u$ to refer to the target $o^*$ under different $\phi_k$, i.e. $P_L(\{o^*, u', o'\}_t\, | \, \phi_k) = S_1(u'_t \,|\, o^*_t, \phi_k)$, and the speaker should likewise use the probability that their partner would produce response $o'$ after hearing utterance $u$, $P_S(\{o^*, u', o'\}_t \,|\, \phi_k) = L_0(o'_t \,|\, u'_t)$,

This symmetry, where each agent is attempting to learn from the other's behavior, creates a clear coordination problem\footnote{In some settings, agents in one role may be expected to take on more of the burden of adaptation, leading to an asymmetric division of labor \cite<e.g.>{MorenoBaggio14_AsymmetrySignaling}. This may be especially relevant in the presence of asymmetries in power, status, or capability \cite{MisyakEtAl14_UnwrittenRules}, but we leave consideration of such asymmetries for future work.}.
In the case of an error, where the agent in the listener role hears the utterance $u'$ and chooses an object $o'$ other than the intended target $o^*$, they will receive feedback about the intended target and subsequently condition on the fact that the speaker chose $u'$ to convey that target.
Meanwhile, the agent in the speaker role will subsequently condition on the likelihood that the listener chose the object $o'$ upon hearing their utterance.
In other words, each agent will subsequently condition on slightly different data leading to conflicting beliefs.
Whether or not agents are able to resolve early misunderstandings through further interaction and eventually reach consensus depends on a number of factors.

\paragraph{The role of memory and recency}

One important constraint is imposed by the basic cognitive mechanisms of memory.
It is unrealistic to expect that memory traces of every past interaction in the set of observations $D$ is equally accessible to the agent.
Furthermore, this may be to the agent's advantage.
Without a mechanism by which errors become less accessible, early misunderstandings may interfere with coordination much later in an interaction. 
One possible solution is to privilege more recent outcomes. 
Especially if a partner is assumed to change over time, then older data may provide less reliable cues to their current behavior.
Recency is typically incorporated into Bayesian models with a simple decay term in the likelihood function \cite{anderson2000adaptive,angela2009sequential,fudenberg2014recency,kalm2018visual}.
$$P(D_k | \phi_k) = \prod_{\tau=0}^T \beta^{\tau} P(\{o^*,u',o'\}_{T-\tau}\, |\, \phi_k)$$
where $\tau=0$ indexes the most recent trial $T$ and decay increases further back through time.
This decay term is motivated by the empirical power function of forgetting \cite{wixted1991form}, and can be derived by simply extending our hierarchical model down an additional layer \emph{within} each partner to allow for the possibility that they are using slightly different lexicons at different points in time; assuming a degree of auto-correlation between neighboring time points yields this form of discounting\footnote{While this simple decay model is sufficient for our reference games, it is clearly missing important mechanistic distinctions between working memory and long-term memory; for example, explaining convention formation over longer timescales may require an explicit model of consolidation or source memory. It is also consistent with multiple algorithmic-level mechanisms; for example, decay can be viewed as a form of weighted importance sampling, where more recent observations are preferentially sampled \cite{pearl2010online}, or a process where observations have some probability of dropping out of memory at each time step.}.

\paragraph{The role of pragmatics}

While natural languages are rife with ambiguous and polysemous terms, speaker and listeners must somehow resolve these ambiguities to be understood in context \cite{PiantadosiTilyGibson12_Ambiguity}\footnote{Indeed, \citeA{brochhagen2020signalling} has suggested that high degrees of lexical ambiguity and polysemy, i.e. high degrees of uncertainty over $\Theta$ in CHAI, are useful precisely because they allow much-needed flexibility supporting partner-specific adaptation.}. 
For example, \citeA{BrennanClark96_ConceptualPactsConversation} placed participants in a context where the target object $o^*$ was easily distinguished from other objects in the context $\mathcal{C}$ by a referring expression like $u=$``the shoe.''
In a second phase of the study, the context $\mathcal{C}'$ was switched to be a set of other shoes.
Even though there was strong precedent for referring to $o^*$ as ``the shoe,'' this description was no longer \emph{informative}: the speaker recognized that $u$ could apply equally well to all $o\in\mathcal{C}$ leading to potential ambiguity about which shoe they were referring to.
As a result, the speaker switched to a more specific utterance like $u'=$``the pennyloafer'' which unambiguously applied to $o^*$ in the new context.
In a third and final phase, the context reverted back to the original one, $\mathcal{C}$, but many speakers continued to use the more specific utterance $u'$ even though $u$ would have sufficed.
This example emphasizes how \emph{ad hoc} conventions or pacts may be sensitive to the context in which they form.

CHAI solves this problem by the principles of \textit{pragmatic reasoning} naturally instantiated in the RSA framework \cite{FrankGoodman12_PragmaticReasoningLanguageGames}, which plays two distinct roles.
First, our Gricean agents'  production and comprehension is guided by cooperative principles (Eq.~\ref{eq:marginalized}).
They do not only make passive inferences from observation, they participate in the interaction by \emph{using language} themselves.
Second, our agents assume that their \emph{partner} is also using language in a cooperative manner, which strengthens the inferences they may make about the underlying system of meanings their partner is using.
That is, we use the RSA equations as the linking function in the likelihood $P(D_k | \phi_k)$, representing an agent's prediction about how a partner with meaning function $\phi_k$ would actually behave in context (Eq.~\ref{eq:joint_inference}). 
This use of pragmatic reasoning has been explicitly linked to principles like mutual exclusivity in word learning \cite{bloom2002children,FrankGoodmanTenenbaum09_Wurwur,SmithGoodmanFrank13_RecursivePragmaticReasoningNIPS,gulordava2020one,ohmerreinforcement}.
For example, upon hearing their partner use a particular utterance $u$ to refer to an object $o$, a pragmatic listener can not only infer that $u$ means $o$ in their partner's lexicon, but also that other utterances $u'$ likely do \emph{not} mean $o$: if they did, the speaker would have used them instead.

\subsection{Simulation details}

While our simulations in the remainder of the paper each address different scenarios, we have aimed to hold as many details as possible constant throughout the paper.
First, we must be concrete about the space of possible lexicons that parameterizes the lexical meaning function, $\mathcal{L}_{\phi}$.
For consistency with previous models of word learning \cite<e.g.>{XuTenenbaum07_WordLearningBayesian} we take the space of possible meanings for an utterance to be the set of nodes in a concept taxonomy.
When targets of reference are conceptually distinct, as typically assumed in signaling games, the target space of utterance meanings reduces to the discrete space of individual objects, i.e.~$\den{u}_{\phi} = \phi(u) \in \mathcal{O}$ for all $u\in\mathcal{U}$.
For this special case, the parameter space contains exactly $|\mathcal{O}| \times |\mathcal{U}|$ possible values for $\phi$, corresponding to all possible mappings between utterances and individual objects. 
Each possible lexicon can therefore be written as a binary matrix where the rows correspond to utterances, and each row contains one object.
The truth-conditional function $\mathcal{L}_{\phi}(u,o)$ then simply checks whether the element in row $u$ matches object $o$.
For example, consider a simple reference game with two utterances and two objects ($o_1=\bluesquare$ and $o_2=\orangecircle$).
Then there are four possible lexicons, corresponding to the four assignments of objects to utterances:
$$\phi \in \left\{\begin{bmatrix}
\bluesquare \\
\orangecircle \\
\end{bmatrix},
\begin{bmatrix}
\orangecircle \\
\bluesquare \\
\end{bmatrix},
\begin{bmatrix}
\orangecircle\\
\orangecircle \\
\end{bmatrix},
\begin{bmatrix}
\bluesquare  \\
\bluesquare \\
\end{bmatrix}\right\}$$

Second, having defined the support of the parameter $\phi$, we can then define a lexical prior. 
We consider a partition-based simplicity prior based on the size of the lexicon \cite{FrankGoodmanTenenbaum09_Wurwur,carr2020simplicity}:
$P(\phi) \propto \exp\{-|\phi|\}$, where $|\phi|$ is the number of lexical items. 
Again, for traditional signaling games, this reduces to a uniform prior because all possible lexicons are the same size: $\phi(u_i) \sim \textrm{Unif}(\mathcal{O})$.
We can compactly write distributions over $\phi$ in terms of the same utterance-object matrix, where row $i$ represents the marginal distribution over possible meanings of utterance $u_i$.
For example, the uninformative prior for two utterances and two objects can be written:
$$P(\phi) =  \begin{array}{cccl}
& &  \bluesquare \,\,\,\,\,\,\orangecircle & \\
\begin{bmatrix}
\textrm{Unif}\{\bluesquare,\orangecircle\} \\
\textrm{Unif}\{\bluesquare,\orangecircle\} \\
\end{bmatrix} & = & \begin{bmatrix}
.5 & .5  \\
.5 & .5 \\
\end{bmatrix}& \hspace{-1em} \begin{array}{r} u_1\\u_2\end{array} 
\end{array}$$
This simplicity prior becomes more important for \textbf{P3}, where we consider spaces of referents with more complex conceptual structure. 
A single word may apply to multiple conceptually related referents (e.g. all of the squares) or, conversely, may apply to no referents at all, in which case it is effectively removed from the agent's vocabulary.
In this case, the simplest lexicon is a single word that refers to everything and the most complex lexicon assigns a unique word for each object (see Appendix C for discussion of alternatives.)

Finally, while the probabilistic model we have formulated in this section is theoretically well-motivated and mathematically well-defined, it is challenging to derive predictions from it.
Historically, interactive models like ours are not amenable to closed-form analytical techniques and computationally expensive to study through simulation, likely contributing to the prevalence of simplified heuristics in prior work. 
Our work has been facilitated by recent advances in probabilistic inference techniques that have helped to overcome these obstacles (see Appendix A for further details of our implementation.)

\subsection{Summary}

In this section, we formalized the computational problem facing agents who must communicate in a variable, changing world. 
No static lexicon is appropriate for all partners and situations, requiring them to update on the fly. 
We proposed CHAI, a cognitive model of how people solve this problem through continual adaptation.
CHAI instantiates three core capacities in a hierarchical Bayesian framework: (C1) structured uncertainty over what words mean to different partners, (C2) social inference to back out likely latent systems of meaning from a partner's observable behavior, and (C3) hierarchical induction to generalize to the overall distribution of possible partners. 
In the remainder of the paper, we argue that CHAI provides a new computational foundation for understanding coordination and convention formation, focusing on three empirical phenomena that have posed a challenge for previous accounts: (P1) the increase in communicative efficiency as a function of shared history, (P2) the transfer of partner-specific expectations to communal expectations, and (P3) the influence of communicative context on which conventions eventually form.

\section{Phenomenon \#1:  \emph{Ad hoc} conventions become more efficient}

We begin by considering the phenomenon of increasing efficiency in repeated reference games: speakers use detailed descriptions at the outset but converge to an increasingly compressed shorthand while remaining understandable to their partner.
While this phenomenon has been extensively documented, to the point of serving as a proxy for measuring common ground, it has continued to pose a challenge for models of communication.
 \begin{figure*}
	\centering
    \includegraphics[scale=.6]{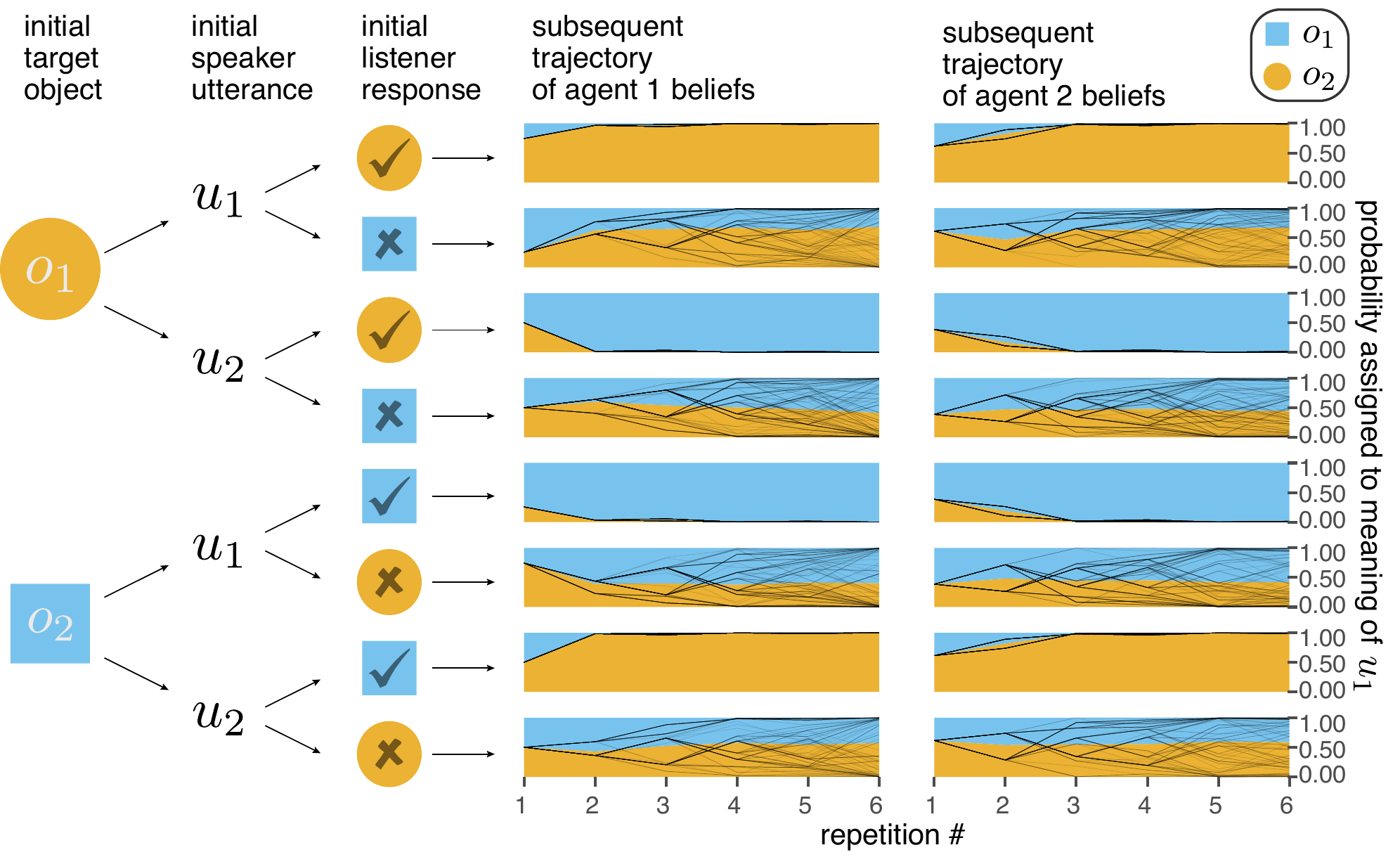}
    \caption{\emph{Path-dependence of conventions.} The average trajectory of each agent's beliefs about the meaning of $u_1$, $\phi(u_1)$, is shown in blue and orange following all eight possible outcomes of the first trial in Simulation 1.1. For each of the two possible targets, the speaker could choose to produce either of the two utterances, and the listener could respond by choosing either of the two objects. In the cases where the listener chose correctly (marked with a checkmark), agents subsequently conditioned on the same data and rapidly converged on a system of meaning consistent with this feedback. For example, in the first row, when $u_1$ was successfully used to refer to the circle, both agents subsequently believe that $u_1$ means \emph{circle} in their partner's lexicon. In the cases where the listener fails to choose the target, the agents subsequently condition on different data, and they converge on a convention that is determined by later choices (lines represent the trajectories of individual agents.)}
  \label{fig:path-dependence}
\end{figure*}
In this section, we argue that CHAI provides a rational explanation for increasing efficiency in terms of the inferences made by speakers across repeated interaction.
Given that this phenomenon arises in purely dyadic settings, it also provides an opportunity to explore more basic properties of the first two capacities formalized in our model (representing \emph{uncertainty} and \emph{partner-specific learning}) before introducing hierarchical generalization in the next section. 

In brief, we show that increasing efficiency is a natural consequence of the speaker's tradeoff between informativity and parsimony (Eq.~\ref{eq:marginalized}), given their inferences about the listener's language model. 
For novel, ambiguous objects like tangrams, where speakers do not expect strong referential conventions to be shared, longer initial descriptions are motivated by high initial uncertainty in the speaker's lexical prior $P(\phi_k | \Theta)$. 
Proposing multiple descriptors is a rational hedge against the possibility that a particular utterance will be misinterpreted and give the listener a false belief.
As the interaction goes on, the speaker obtains feedback $D_k$ from the listener responses and updates their posterior beliefs $P(\phi_k | D_k)$ accordingly. 
As uncertainty gradually decreases, they are able to achieve the same expected informativity with shorter, more efficient messages.

\begin{figure}[t]
\centering
    \includegraphics[scale=.8]{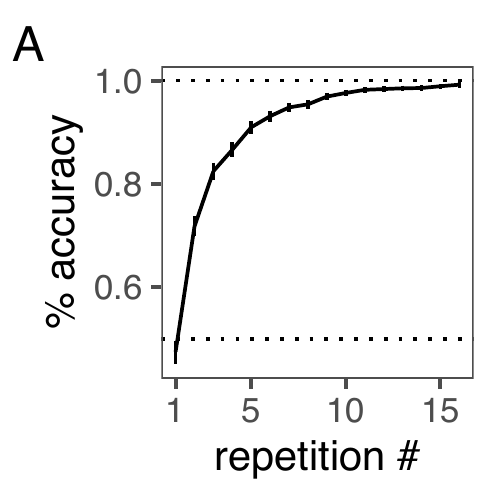}\includegraphics[scale=.8]{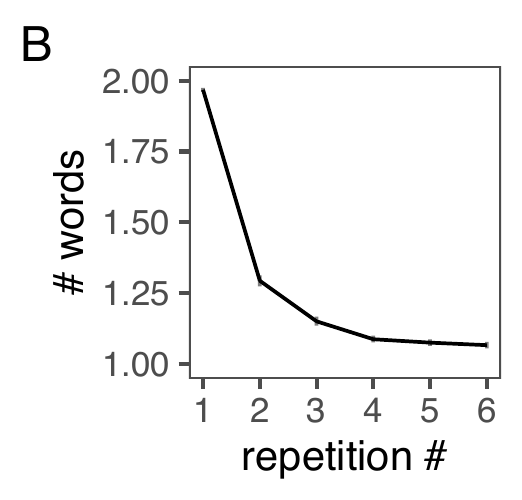}
  \caption{\emph{Pairs of agents learn to successfully coordinate on efficient ad hoc conventions over repeated interactions.} (A) agents converge on accurate communication systems in Simulation 1.1, where only single-word utterances are available, and (B) converge on shorter, more efficient conventions in Simulation 1.2, where multi-word utterances were available. Error bars are bootstrapped 95\% CIs across 1000 trajectories, computed within each repetition block of two trials.}
  \label{fig:sec1model}
\end{figure}

\subsection{Simulation 1.1: Pure coordination}

We build up to our explanation of increasing efficiency by first exploring a traditional signaling game scenario with only one-word utterances.
This simulation tests the most fundamental competency for any model of \emph{ad hoc} coordination: agents are able to coordinate on a communication system in the absence of shared priors. 
We consider the simplest possible reference game with two objects, $\mathcal{O} = \{\orangecircle, \bluesquare\}$, where the speaker must choose between two one-word utterances $\mathcal{U} = \{u_1, u_2\}$ with equal production cost. 

We walk explicitly through the first step of the simulation to illustrate the model's dynamics (see Fig.~\ref{fig:path-dependence}).
Suppose the target object presented to the speaker agent on the initial trial is $\orangecircle$.
Both utterances are equally likely to apply to either object under the uniform lexical prior, hence each utterance is expected to be equally (un)informative. 
The speaker's utility therefore reduces to sampling an utterance at random $u~\sim~S(u\,|\, \orangecircle)$.
Suppose $u_1$ is sampled.
The listener then hears this utterance and selects an object according to their own expected utility under their uniform lexical prior, which also reduces to sampling an object at random $o \sim L(o | u_1)$.
Suppose they choose, $ \orangecircle$, a correct response.
Both agents may use the resulting tuple $D = \{ \orangecircle^*, u_1,  \orangecircle\}$, depicted in the top row in Fig.~\ref{fig:path-dependence} to update their beliefs about the lexicon their partner is using.
\begin{align}
P_S(\phi | D) & \propto L_0( \orangecircle\, |\, u_1, \phi)P(\phi)\nonumber \\
P_L(\phi | D) & \propto S_1(u_1 \,|\, \orangecircle^*, \phi)P(\phi)\nonumber
\end{align}
They then proceed to the next trial, where they use this updated posterior distribution to produce or interpret language instead of their prior.
To examine how the dynamics of this updating process unfold over further rounds, we simulated 1000 such trajectories.
The trial sequence was structured as a repeated reference game, containing 30 trials structured into 15 repetition blocks.
The two objects appeared in a random order within each block, and agents swapped roles at the beginning of each block.
We show representative behavior at soft-max optimality parameter values $\alpha_L = \alpha_S = 8$ and memory discounting parameter $\beta = 0.8$, but find similar behavior in a wide regime of parameter values (see Appendix Fig.~\ref{fig:arbitrariness_grid}).

We highlight several key results from this simulation.
First, and most fundamentally, the communicative success of the dyad rises over the course of interaction: the listener is able to more accurately select the intended target object (see Fig.~\ref{fig:sec1model}A). 
Second, the initial symmetry between meanings in the prior is broken by initial choices, leading to \emph{arbitrary} but \emph{stable} mappings in future rounds.
Because agents were initialized with the same priors in every trajectory, trajectories only diverged when different actions happen to be sampled.
This can be seen by examining the path-dependence of subsequent beliefs based on the outcome of the initial trial in Fig.~\ref{fig:path-dependence}.
Third, we observe the influence of mutual exclusivity via Gricean pragmatic reasoning: agents also make inferences about objects and utterances that were \emph{not} chosen. 
For example, observing $D = \{(\orangecircle^*, u_2, \orangecircle)\}$ provides evidence that $u_1$ likely does not mean $\orangecircle$ (e.g.~the third row of Fig.~\ref{fig:path-dependence}, where hearing $u_2$ refer to $\orangecircle$ immediately led to the inference that $u_1$ likely refers to $\bluesquare$).

\subsection{Simulation 1.2: Increasing efficiency}

Next, we show how our model explains speakers' gains in efficiency over multiple interactions. 
For efficiency to change at all, speakers must be able to produce utterances that vary in length. 
For this simulation, we therefore extend the model to allow for multi-word utterances by allowing speakers to combine together multiple primitive utterances.
Intuitively, human speakers form longer initial description by combining a collection of simpler descriptions (e.g. ``kind of an X, or maybe a Y with Z on top''). 
This raises a problem about how the meaning of a multi-word utterance $\mathcal{L}_\phi(u_1u_2)$ is derived from its components $\mathcal{L}_\phi(u_1)$ and $\mathcal{L}_\phi(u_2)$.
To capture the basic desideratum that an object should be more likely to be chosen by $L_0$ when more components of the longer utterance apply to it, we adopt a standard conjunctive semantics:
$$\mathcal{L}_\phi(u_iu_j, o) = \mathcal{L}_\phi(u_i, o) \times \mathcal{L}_\phi(u_j, o)$$
One subtle consequence of a conjunctive Boolean semantics is the possibility of contradictions. 
For example, under a possible lexicon where $\phi(u_1)=\bluesquare$ and $\phi(u_2)=\orangecircle$, the multi-word utterance $u_1u_2$ is not only false of the particular referents in the current context, it is false of all \emph{possible} referents, reflecting a so-called truth-gap \cite{Strawson50_OnReferring,van1966singular}. 
We assume such an utterance is uninterpretable and simply disregarded without changing the literal listener $L_0$'s beliefs. 
While this assumption is sufficient for our simulations, we regard this additional complexity as a limitation of classical truth-conditional semantics  \cite{degen2020redundancy} and show in Appendix B that switching to a continuous semantics with lexical values in the interval $[0,1]$ may better capture the notion of redundancy that motivates speakers to initially produce longer utterances.

\begin{figure*}[t]
\centering
    \includegraphics[scale=.9]{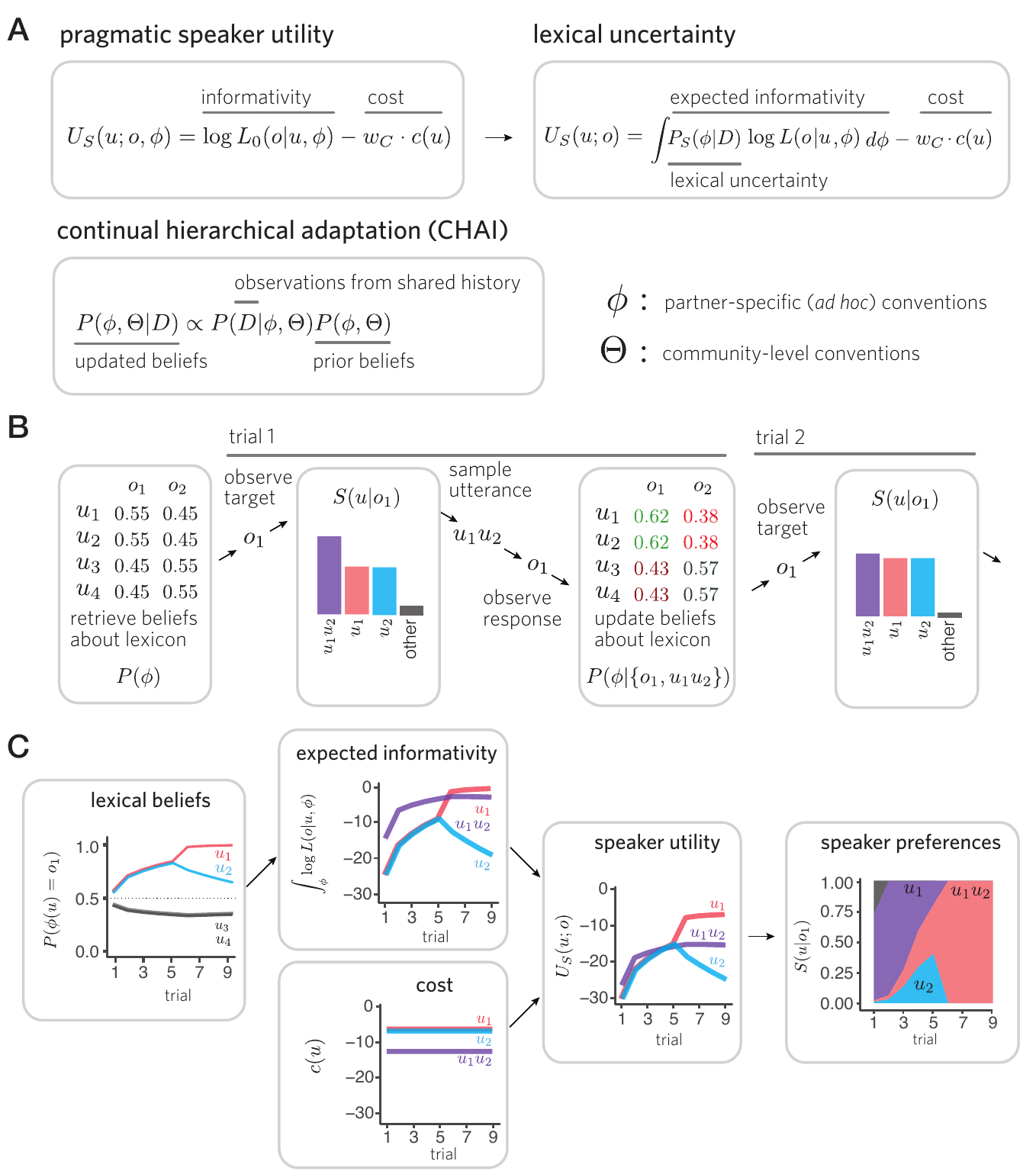}
  \caption{\emph{Internal state of speaker.} (A) Schematic showing how lexical uncertainty is added to a simple pragmatic speaker utility; CHAI proposes that lexical expectations are adapted over time based on social observations. (B) A single trial of Simulation 2.1. The speaker begins with uncertainty about the meanings in the listener's lexicon (e.g. assigning 55\% probability to the possibility that utterance $u_1$ means object $o_1$.) A target $o_1$ is presented, and the speaker samples an utterance from the distribution $S(u|o_1)$. Finally, they observe the listener's response and update their beliefs. Due to the compositional semantics of the utterance $u_1u_2$, the speaker becomes increasingly confident that both component primitives, $u_1$ and $u_2$, apply to object $o_1$ in their partner's lexicon. (C) Each internal term of the speaker's utility (Eq. \ref{eq:marginalized}) is shown throughout the interaction. When the speaker is initially uncertain about meanings (far left), the longer utterance $u_1u_2$ has higher expected informativity (center-left) and therefore higher utility (center-right) than the shorter utterances $u_1$ and $u_2$, despite its higher cost (far-right). As the speaker observes several successful interactions, they update their beliefs and become more confident about the meanings of the component lexical items $u_1$ and $u_2$. As a result, more efficient single-word utterances gradually gain in utility as cost begins to dominate the utility. On trial 5, $u_1$ is sampled, breaking the symmetry between utterances.}
  \label{fig:sec1internals}
\end{figure*}

Now, we consider a scenario with the same two objects as in Simulation 1.1, but give the speaker four primitive utterances $\{u_1, u_2, u_3, u_4\}$ instead of only two, and allow two-word utterances such as $u_1u_2$.
We established in the previous section that successful \emph{ad hoc} conventions can emerge even in a state of pure uncertainty, but human participants in repeated reference games typically bring some prior expectations about language into the interaction.
For example, a participant who hears `ice skater' on the first round of the task in \citeA{ClarkWilkesGibbs86_ReferringCollaborative} may be more likely to select some objects more than others while still having substantial uncertainty about the intended target (e.g. over three of the twelve tangram that have some resemblance to an ice skater).
We thus initialize both agents with weak biases $\delta$ (represented in compressed matrix form in Fig.~\ref{fig:sec1internals}):
\begin{align}
\phi(u_1), \phi(u_2) \sim \textrm{Categorical}(0.5 + \delta) \nonumber\\
\phi(u_3), \phi(u_4) \sim \textrm{Categorical}(0.5 - \delta) \nonumber
\end{align}


As in Simulation 1.1, we simulated 1000 distinct trajectories of dyadic interaction between agents.
Utterance cost was defined to be the number of `words' in an utterance, so $c(u_1) =1$ and $c(u_1u_2)=2$.
As shown in Fig.~\ref{fig:sec1model}B, our speaker agent initially prefers longer utterance (mean length $\approx 1.5$ on first block) but rapidly converges to shorter utterances after several repetitions (mean length $\approx 1$ on final block), qualitatively matching the curves measured in the empirical literature.

To illustrate in detail how our model derives this behavior, we walk step-by-step through a single trial (Fig. \ref{fig:sec1internals}).
Consider a speaker who wants to refer to object $\bluesquare$. 
They expect their partner to be slightly more likely to interpret their language using a lexicon in which $u_{1}$ and $u_{2}$ apply to this object, due to their weak initial biases. 
However, there is still a reasonable chance ($p=0.45$) that either $u_1$ or $u_2$ alone will be interpreted to mean $\orangecircle$, giving their partner false beliefs. 
To see why our speaker model initially prefers the longer utterance $u_{1}u_{2}$ to hedge against this possibility, despite its higher production cost, consider the expected informativity of $u_1u_2$ under different possible lexicons.
The possibility with highest probability is that both $\phi(u_1) = \phi(u_2) = \bluesquare$ in the listener's lexicon ($p = 0.55^2 \approx 0.3$), in which case the listener will correctly identify $\bluesquare$ with high probability.
The possibility that both $\phi(u_1)=\phi(u_2) = \orangecircle$ in the listener's lexicon is only $p=0.45^2 \approx 0.2$, in which case the listener will erroneously select $\orangecircle$.
In the mixed cases, where $\phi(u_1) = \orangecircle, \phi(u_2) = \bluesquare$ or $\phi(u_1) = \bluesquare, \phi(u_2) = \orangecircle$ in the listener's lexicon ($p = 2 \cdot 0.45 * 0.55 \approx 0.5$), the utterance would be a interpreted as a contradiction and the listener would not change their prior beliefs.
Because the speaker's informativity is defined using the log probability of the listener's belief, the utility of giving the listener a false belief, $\log(\epsilon)$ is significantly worse than simply being uninformative, i.e. $\log(0.5)$, and the longer utterance minimizes this harm.

Following the production of a conjunction, the speaker observes the listener's response (say, $\bluesquare$).
This allows both agents to become more confident that the component utterances $u_1$ and $u_2$ mean $\bluesquare$ in their updated posterior over the listener's lexicon.
This credit assignment to individual lexical items is a consequence of the compositional meaning of longer utterances in our simple grammar.
The listener knows a speaker for whom either $u_1$ or $u_2$ individually means $\bluesquare$ would have been more likely to say $u_1u_2$ than a speaker for whom either component meant $\orangecircle$; and similarly for the speaker reasoning about possible listeners.
Consequently, the probability of both mappings increases.

Fig.~\ref{fig:sec1internals} shows the trajectories of internal components of the speaker utility as the interaction continues.
We assume for illustrative purposes in this example that $\bluesquare$ continues to be the target on each trial and the same agent continues to be the speaker.
As the posterior probability that individual primitive utterances $u_1$ and $u_2$ independently mean $\bluesquare$ increases (far left), the marginal gap in informativity between the conjunction and the shorter components gradually decreases (center left).
As a consequence, production cost increasingly dominates the utility (center-right). 
After several trials of observing a successful listener response given the conjunction, the \emph{informativity} of the two shorter utterances reaches parity with the conjunction but the cost makes the shorter utterances more attractive (yielding a situation now similar to the outset of Simulation 1.1).
Once the speaker samples one of the shorter utterances (e.g. $u_1$), the symmetry collapses and that utterance remains most probable in future rounds, allowing for a stable and efficient \emph{ad hoc} convention.
Thus, increasing efficiency is derived as a rational consequence of uncertainty and partner-specific inference about the listener's lexicon.
For these simulations, we used $\alpha_S = \alpha_L = 8, w_c = 0.24, \beta=0.8$ but the qualitative reduction effect is found over a range of different parameters (see Appendix Fig. \ref{fig:conjunction_grid}). 

%
%
%


\subsection{Discussion}

The simulations presented in this section aimed to establish a rational explanation for feedback-sensitive increases in efficiency over the course of \emph{ad hoc} convention formation.
Speakers initially hedge their descriptions under uncertainty about the lexical meanings their partner is using, but are able to get away with less costly components of those descriptions as their uncertainty decreases.
This explanation recalls classic observations about \emph{hedges} (expressions like \emph{sort of} or morphemes like \emph{-ish}) that explicitly mark provisionality, such as \emph{a sort of silvery purple colored car} \cite{lakoff1975hedges,Fraser10_Hedging,MedlockBriscoe07_HedgeClassification}.
\citeA{BrennanClark96_ConceptualPactsConversation} counted hedges across repetitions of a repeated reference game, finding a greater occurrence of hedges on early trials than later trials and a greater occurrence under more ambiguous contexts.
While our model does not include hedges, it is possible to understand this behavior as an explicit or implicit marker of the lexical uncertainty construct in our account.
Our account is also broadly consistent with recent analyses of exactly \emph{what} gets reduced in a large corpus of repeated reference games \cite{hawkins2020characterizing}.
These analyses found that entire modifying clauses are more likely to be dropped at once than would be expected by random and independent corruption.
In other words, speakers apparently begin by combining multiple descriptive modifiers and collapse to retain only one of these `units' contingent on evidence that their partner understands.

Why has this phenomenon remained outside the explanatory scope of previous models?
Our account differs in both level of analysis and model complexity.
For example, the influential \emph{interactive alignment} account proposes that that speakers adapt and coordinate on meaning through priming mechanisms that allow phonetic or syntactic features associated with lexical items to percolate up to strengthen higher levels of representation \cite{pickering2004toward, pickering2006alignment,roelofs1992spreading}.
While priming mechanisms are certainly at play in repeated reference tasks, especially when listeners engage in extensive dialogue and alternate roles, it is not clear why priming alone would lead to convergence on more efficient descriptions as opposed to aligning on the same longer initial description.
Furthermore, priming cannot explain why speakers still converge to shorter descriptions even when the listener is prevented from saying anything at all and only sparse, non-verbal feedback of success is provided, or why speakers continue using longer descriptions when they receive non-verbal feedback that the listener is repeatedly making errors (\citeNP{KraussWeinheimer66_Tangrams}; see also \citeNP{hawkins2020characterizing}).
In these cases, there are no linguistic features available for priming or alignment to act upon.
To be clear, our computational-level account is not mutually exclusive with these process-level principles and does not in any way falsify or undermine them.
Explaining when and why speakers believe that shorter descriptions will suffice, and how it depends on context, requires additional computational-level principles, which we hope will lead to further enrichment of algorithms at the process level.

Another prominent account proposes that speakers coordinate on meaning using a simpler update rule that simpler makes utterances more likely to be produced after communicative successes and less likely after communicative failures.
This account has often been implemented using a simple variant of reinforcement learning (RL) such as Roth-Erev learning \cite{erev1998predicting,steels_self-organizing_1995,barr_establishing_2004,young_evolution_2015}.
While such minimal rules allow groups to reach consensus, it is challenging to explain the full suite of phenomena we have explored in this section. 
First, it is not clear how simply reinforcing longer descriptions could lead them to get shorter. 
In the rare cases that have allowed longer utterances to be constructed compositionally from more primitive utterances, reduction has been hard-coded as a kind of $\epsilon$-greedy exploration where the speaker has a fixed probability of dropping a random token at each point in time \cite{beuls2013agent,steels2016agent}.
Such noisy dropping, however, is inconsistent with studies by \citeA{HupetChantraine92_CollaborationOrRepitition} where participants were asked to repeatedly refer to the same targets for a \emph{hypothetical} partner to see later, such that any effects of familiarity or repetition on the part of the speaker would be the same as the interactive task.
No evidence of reduction was found in this case, and in some cases utterances actually grew longer \cite<see also>{GarrodFayLeeOberlanderMacLeod07_GraphicalSymbolSystems}.
Even if we fixed this problem by extending the update rule to be contingent on interaction, it is not clear why a speaker would initially prefer to produce longer utterances over shorter utterances.

Importantly, these limitations do not stem from the RL framework itself, but from the simplifying assumption that the probability of taking actions should be directly tied to the previous outcomes of those actions.
CHAI preserves a core idea from these accounts --- the ability to dynamically adapt one's behavior contingent on one's partner's --- but disentangles the inference problem (i.e. estimating a partner's underlying lexicon) from the decision problem (i.e. deciding which action to take with these estimates in hand).
Introducing the latent variable of the lexicon increases the model's complexity but is also more explanatory, as we show in the subsequent sections. 
Importantly, more sophisticated model-based reinforcement learning algorithms make a similar distinction and may consequently be flexible enough to account for this phenomenon (see \citeNP{gershman2015novelty} for an explicit connection between hierarchical Bayes and an RL algorithm known as TD-learning; but see \citeNP{velez2021learning} for outstanding problems associated with bridging these perspectives).

Finally, while our simulations captured several core features of the reduction phenomenon, they have only scratched the surface of its empirical complexity.
First, our simulations only consider two-word descriptions with homogenous uncertainty over the components, while the semantic components of real initial descriptions have more heterogeneity. 
It remains an open question as to how best to instantiate more realistic priors in our model that can predict more fine-grained patterns. 
For example, early hand-tagged analyses by \citeA{Carroll80_NamingHedges} found that in three-quarters of transcripts from \citeA{krauss_changes_1964} the conventions that participants eventually converged upon were prominent in some syntactic construction at the beginning, often as a head noun that was initially modified or qualified by other information. 
Second, gains in efficiency associated with \emph{ad hoc} conventions do not necessarily translate into shorter utterances.
Outside of the domain of reference games, speakers often have control over what they want to convey and may use the efficiency afforded by their new conventions to express more information in the same number of words rather than the same amount of information in fewer words \cite{effenberger2021analysis}. 
Once a convention is formed, it can be used as a new primitive to bootstrap further conventions and convey ever-more-sophisticated meanings \cite{mccarthy2021learning}.

\section{Phenomenon \#2:  Conventions gradually generalize to new partners in community}

How do we make the inferential leap from \emph{ad hoc} conventions formed through interaction with a single partner to \emph{global} conventions expected to be shared throughout a community?
Grounding collective convention formation in the individual learning mechanisms explored in the previous section requires an explicit \emph{theory of generalization} capturing how people transfer what they have learned from one partner to the next.
One influential theory is that speakers simply ignore the identity of different partners and update a single monolithic representation after every interaction \cite{steels_self-organizing_1995,barr_establishing_2004,young_evolution_2015}.
We call this a \emph{complete-pooling} theory because data from each partner is collapsed into an undifferentiated pool of evidence \cite{gelman2006data}. 
Complete-pooling models have been remarkably successful at predicting collective behavior on networks, but have typically been evaluated only in settings where anonymity is enforced. 
For example, \citeA{centola_spontaneous_2015} asked how large networks of participants coordinated on conventional names for novel faces.
On each trial, participants were paired with a random neighbor but were not informed of that neighbor's identity, or the total number of different possible neighbors. 

While complete-pooling may be appropriate for some everyday social interactions, such as coordinating with anonymous drivers on the highway, it is less tenable for everyday communicative settings.
Knowledge about a partner's identity is both available and relevant for conversation \cite{eckert_three_2012, davidson_nice_1986}.
Partner-specificity thus poses clear problems for complete-pooling theories but can be easily explained by another simple model, where agents maintain separate expectations about meaning for each partner.
We call this a \emph{no-pooling} model \cite<see>[which contrasted no-pooling and complete-pooling models]{SmithEtAl17_LanguageLearning}.
The problem with no-pooling is that agents are forced to start from scratch with each partner.
Community-level expectations never get off the ground.

In other words, complete-pooling and no-pooling models are \emph{prima facie} unable to explain partner-specificity and network convergence, respectively. 
CHAI is a hierarchical \emph{partial-pooling} account that offers a solution to this puzzle. 
We propose that social beliefs about language have hierarchical structure.
That is, the meanings used by different partners are expected to be drawn from a shared community-wide distribution but are also allowed to differ from one another in systematic, partner-specific ways.
This structure provides an inductive pathway for abstract population-level expectations to be distilled from partner-specific experience.
The key predictions distinguishing our model thus concern the pattern of generalization across partners.
Experience with a single partner ought to be relatively uninformative about further partners, hence our partial-pooling account behaves much like a no-pooling model in predicting strong partner-specificity and discounting outliers \cite<see>[which explores this prediction in a developmental context]{dautriche2021}.
After interacting with multiple partners in a tight-knit community, however, speakers should become increasingly confident that labels are not simply idiosyncratic features of a particular partner's lexicon but are shared across the entire community, gradually transitioning to the behavior of a complete-pooling model.
In this section, we test this novel prediction in a networked communication game.
We then explicitly compare CHAI to complete-pooling and no-pooling sub-models that lesion the hierarchy, using only the top level or bottom level, to evaluate the contribution of each component.

\subsection{Model predictions: Simulation 2.1}

\begin{figure}[t]
\centering
\includegraphics[scale=.8]{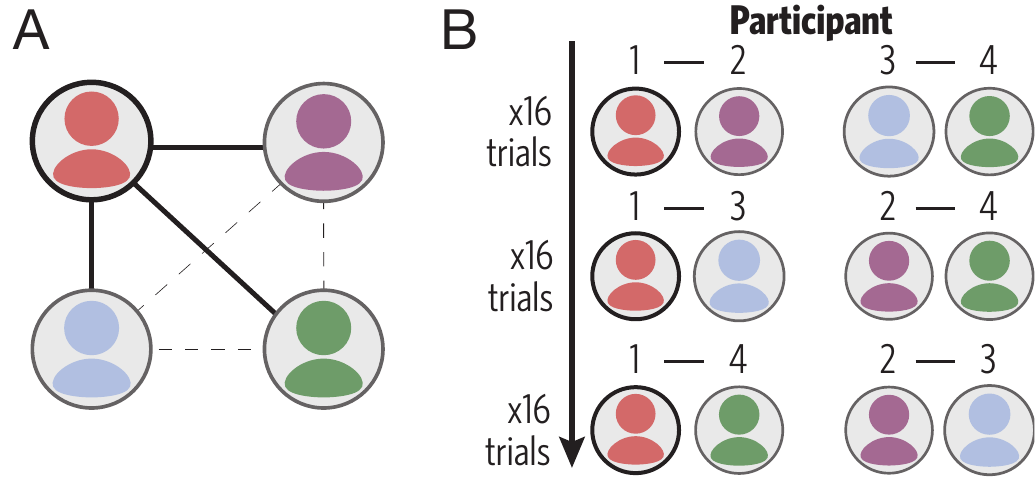}
\vspace{.1em}
\caption{In our simulations and behavioral experiment, participants were  (A) placed in fully-connected networks of 4, and (B) paired in a round-robin schedule of repeated reference games with each neighbor.}
\label{fig:task1_display}
\end{figure}

We first examine the generalization behavior produced by each model by simulating the outcomes of interacting with multiple partners on a small network (see Fig.~\ref{fig:task1_display}A). 
We used a round-robin scheme (Fig.~\ref{fig:task1_display}B) to schedule four agents into a series of repeated reference games with their three neighbors, playing 8 successive trials with one partner before advancing to the next, for a total of 24 trials.
These reference games used a set of two objects $\{o_1, o_2\}$ and four utterances $\{u_1, u_2, u_3, u_4\}$ as in Simulation 1.2; agents were randomized to roles when assigned to a new partner and swap roles after each repetition block within a given interaction.
Consequently, all agents at a particular phase have interacted with the same number of previous partners, allowing us to examine network convergence \cite<but see>[for a ``first-person'' version where each new partner is entirely fresh to the task, finding similar speaker generalization]{hawkins2020generalizing}.

Unlike our previous simulations with a single partner, where hierarchical generalization was irrelevant, we must now specify the hyper-prior P($\Theta$) governing the overall distribution of \emph{partners} (Eq.~\ref{eq:joint_inference}).
Following \citeA{KempPerforsTenenbaum07_HBM}, we extend the uniform categorical prior over possible referents to a hierarchical Dirichlet-Multinomial model \cite{gelman_bayesian_2014}, where the prior over the partner-specific meaning of $u$, $P(\phi_k(u_i)~=~o_j)$, is not uniform, but given by a parameter $\Theta$  that is shared across the entire population.
Because $\Theta$ is a vector of probabilities that must sum to 1 across referents, we assume it is drawn from a Dirichlet prior:
\begin{equation}
\label{eq:top_level}
\begin{array}{rcl}
\phi_k(u)  &\sim&  \textrm{Categorical}(\Theta) \\
\Theta & \sim & \textrm{Dirichlet}(\lambda \cdot \boldsymbol{\alpha})
\end{array}
\end{equation}
where $\lambda \cdot \boldsymbol{\alpha}$ gives the concentration parameter encoding the agent's beliefs, or ``over-hypotheses'' about both the central tendency and the variability of lexicons in the population. 
The relative values of the entries of $\boldsymbol{\alpha}$ correspond to inductive biases regarding the central tendency of lexicons, while the absolute magnitude of  the scaling factor $\lambda$ roughly corresponds to prior beliefs about the spread, where larger magnitudes correspond to more concentrated probability distributions across the population.
We fix $\lambda = 2$ and assume the agent has uncertainty about the population-level central tendency by placing a hyper-prior on $\alpha$ \cite<see>{cowans2004information} that roughly corresponds to the weak initial preferences we used in our previous simulations:
\begin{figure*}[t]
\centering
\includegraphics[scale=.8]{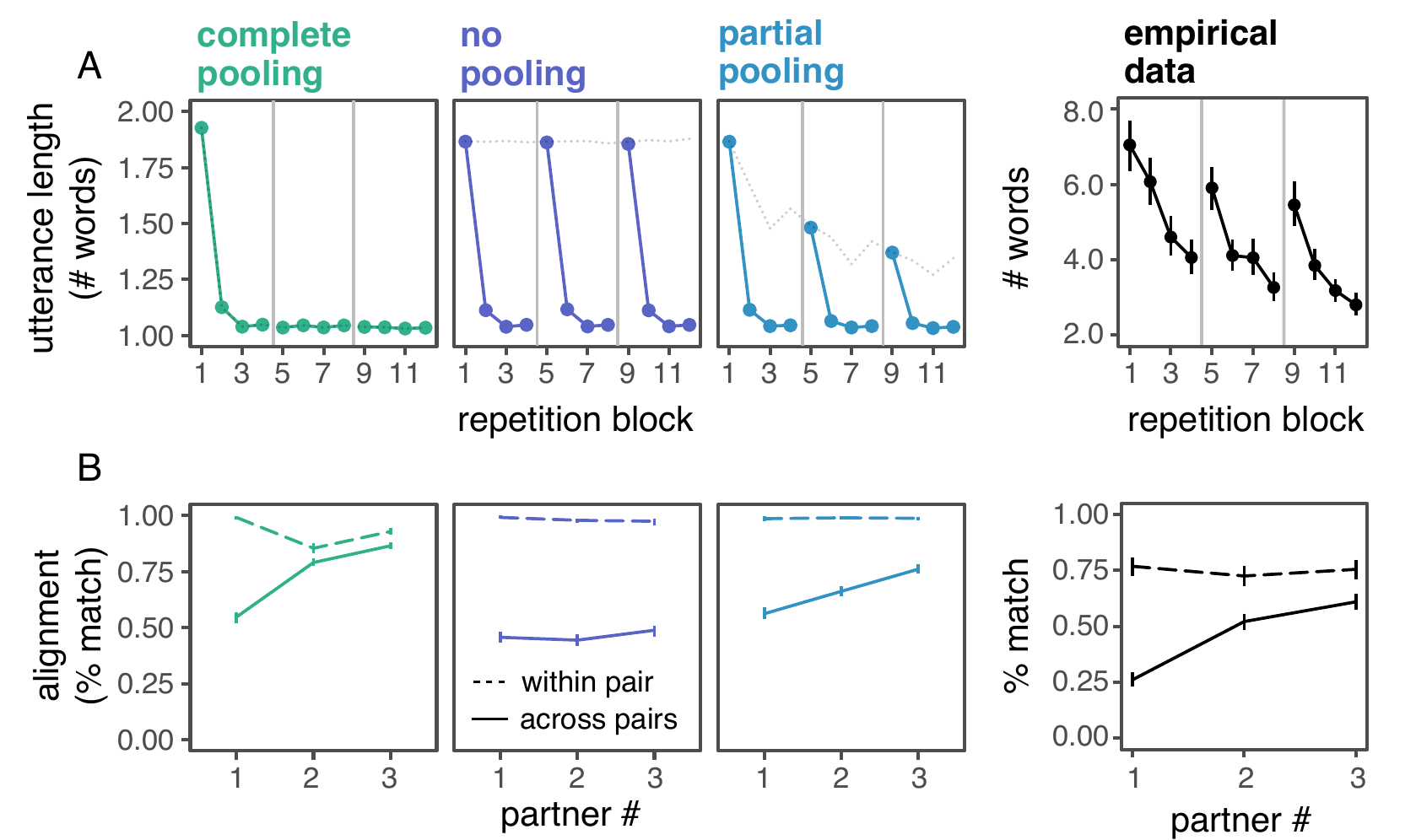}
\vspace{1em}
\caption{Simulation results and empirical data for (A) speaker reduction, and (B) network convergence across three partners. In (A) vertical boundaries mark time points when new partners were introduced, and the dotted grey line represents what would be produced for a stranger at each point in time. In (B), dashed line represents alignment between partners who are currently paired while solid line represents alignment across partners who are \emph{not} paired. Error bars represent bootstrapped 95\% confidence intervals.}
\label{fig:results}
\end{figure*}

\begin{equation*}
\begin{array}{rcl}
 \boldsymbol{\alpha} & \sim& \left\{
\begin{array}{rcl}
  \textrm{Dirichlet}(1.0,1.5)& \textrm{if} & u \in \{u_1, u_2\}\\
  \textrm{Dirichlet}(1.5,1.0) & \textrm{if} &u \in \{u_3, u_4\}\\
\end{array}\right. 
\end{array}
\end{equation*}

We may then define the no-pooling and complete-pooling models by lesioning this shared structure in different ways. 
The no-pooling model assumes an independent $\Theta_k$ for every partner, rather than sharing a single population-level parameter.
Conversely, the complete-pooling model assumes a single, shared $\phi$ rather than allowing different values $\phi_k$ for different partners.
We simulated 48 networks for each model, setting $\alpha_S = \alpha_L = 4,~w_C = .24$ (see Fig.~\ref{fig:partnerspecificity_grid} in the Appendix for an exploration of other parameters).

\paragraph{Speaker utterance length across partners}

We begin by examining our model's predictions about how a speaker's referring expressions change with successive listeners.
While it has been frequently observed that messages reduce in length across repetitions with a single partner \cite{krauss_changes_1964} and sharply revert back to longer utterances when a new partner is introduced \cite{wilkes-gibbs_coordinating_1992}, the key prediction distinguishing our model concerns behavior across subsequent partner boundaries.
Complete-pooling accounts predict no reversion in number of words when a new partner is introduced  (Fig.~\ref{fig:results}A, first column).
No-pooling accounts predict that roughly the same initial description length will re-occur with every subsequent interlocutor  (Fig.~\ref{fig:results}A, second column). 

Here we show that a partial pooling account predicts a more complex pattern of generalization.
First, unlike the complete-pooling model, we find that the partial-pooling speaker model reverts or jumps back to a longer description at the first partner swap.
This reversion is due to ambiguity about whether the behavior of the first partner was idiosyncratic or attributable to community-level conventions.
In the absence of data from other partners, a partner-specific explanation is more parsimonious.
Second, unlike a no-pooling model, after interacting with several partners, the model becomes more confident that one of the short labels is shared across the entire community, and is correspondingly more likely to begin a new interaction with it (Fig.~\ref{fig:results}A, third column).

It is possible, however, that these two predictions only distinguish our partial-pooling model at a few parameter values; the no-pooling and complete-pooling could produce these qualitative effects elsewhere in parameter space.
To conduct a more systematic model comparison, then, we simulated 10 networks in each cell of a large grid manipulating the the optimality parameters $\alpha_S,\alpha_L$, the cost parameter $w_C$, and the memory discounting parameter $\beta$. 
We computed a ``reversion'' statistic (the magnitude of the change in $P(u_1u_2)$ immediately after a partner swap) and a ``generalization'' statistic (the magnitude of the change in $P(u_1u_2)$ from the initial trial with the agent's first partner to the initial trial with the final partner) and conducted single-sample $t$-tests at each parameter value to compare these statistics with what would be expected due to random variation.
We found that only the partial-pooling model consistently makes both predictions across a broad regime. 
The complete-pooling model fails to predict reversion nearly everywhere while the no-pooling model fails to predict generalization nearly everywhere.
Detailed results are shown in Fig.~\ref{fig:generalization_modelcomparison} in the Appendix.

\paragraph{Network convergence}

Because all agents are simultaneously making inferences about the others, the network as a whole faces a coordination problem.
For example, in the first block, agents 1 and 2 may coordinate on using $u_1$ to refer to $o_1$ while agent 3 and 4 coordinate on using $u_2$. 
Once they swap partners, they must negotiate this potential mismatch in usage. 
How does the network as a whole manage to coordinate?
We measured alignment by examining the intersection of utterances produced by speakers: if two agents produced overlapping utterances to refer to a given target (i.e.~a non-empty intersection), we assign a 1, otherwise we assign a 0.
We calculated alignment between currently interacting agents (i.e. \emph{within} a dyad) and those who were not interacting (i.e. \emph{across} dyads), averaging across the target objects.
Alignment across dyads was initially near chance, reflecting the arbitrariness of whether speakers reduce to $u_1$ or $u_2$. 
Under a complete-pooling model (Fig.~\ref{fig:results}B, first column), agents sometimes persist with mis-calibrated expectations learned from previous partners rather than adapting to their new partner, and \emph{within-dyad} alignment deteriorates, reflected by a sharp drop from 99\% to 85\%.
Under a no-pooling model (Fig.~\ref{fig:results}B, second column), convergence on subsequent blocks remains near chance, as conventions need to be re-negotiated from scratch.
By contrast, under our partial-pooling model, alignment across dyads increases without affecting alignment within dyads, suggesting that hierarchical inference leads to emergent consensus (Fig.~\ref{fig:results}B, third column).

\subsection{Behavioral experiment}

To evaluate the predictions derived in our simulations, we designed a natural-language communication experiment following roughly the same network design as our simulations.
That is, instead of anonymizing partners, as in many previous empirical studies of convention formation \cite<e.g.>{centola_spontaneous_2015}, we divided the experiment into blocks of extended dyadic interactions with stable, identifiable partners \cite<see>[for similar designs]{fay_interactive_2010, garrod_conversation_1994}.
Each block was a full repeated reference game, where participants had to coordinate on \emph{ad hoc} conventions for how to refer to novel objects with their partner.
Our partial-pooling model predicted that these conventions will partially reset at partner boundaries, but agents should be increasingly willing to transfer expectations from one partner to another.

\paragraph{Participants}

We recruited 92 participants from Amazon Mechanical Turk to play a series of interactive, natural-language reference games using the framework described in \citeA{Hawkins15_RealTimeWebExperiments}.

\paragraph{Stimuli and procedure}

Each participant was randomly assigned to one of 23 fully-connected networks with three other participants as their neighbors (Fig. \ref{fig:task1_display}A). 
Each network was then randomly assigned one of three distinct contexts containing abstract tangram stimuli taken from \cite{ClarkWilkesGibbs86_ReferringCollaborative}.
The experiment was structured into a series of three repeated reference games with different partners, using these same four stimuli as referents.
Partner pairings were determined by a round-robin schedule (Fig. \ref{fig:task1_display}B).
The trial sequence for each reference game was composed of four repetition blocks, where each target appeared once per block.
Participants were randomly assigned to speaker and listener roles and swapped roles on each block.
After completing sixteen trials with one partner, participants were introduced to their next partner and asked to play the game again. 
This process repeated until each participant had partnered with all three neighbors.
Because some pairs within the network took longer than others, we sent participants to a temporary waiting room if their next partner was not ready. 

Each trial proceeded as follows.
First, one of the four tangrams in the context was highlighted as the \emph{target object} for the speaker.
They were instructed to use a chatbox to communicate the identity of this object to their partner, the listener.
The two participants could engage freely in dialogue through the chatbox but the listener must ultimately make a selection from the array. 
Finally, both participants in a pair were given full feedback on each trial about their partner's choice and received bonus payment for each correct response. 
The order of the stimuli on the screen was randomized on every trial to prevent the use of spatial cues (e.g. ``the one on the left'').
The display also contained an avatar for the current partner  representing different partners with different colors as shown in Fig.~\ref{fig:task1_display} to emphasize that they were speaking to the same partner for an extended period.
On the waiting screen between partners, participants were shown the avatars of their previous partner and upcoming partner and told that they were about to interact with a new partner.

\subsection{Results}

We evaluated participants' generalization behavior on the same metrics we used in our simulations: utterance length and network convergence.

\paragraph{Speaker utterance length}

Now we are in a position to evaluate the central prediction of our model.
Our partial pooling model predicts (1) gains in efficiency within interactions with each partner and (2) reversions to longer utterances at partner boundaries, but (3) gradual shortening of  the initial utterance chosen with successive partners.
As a measure of efficiency, we calculated the raw length (in words) of the utterance produced on each trial.
Because the distribution of utterance lengths is heavy-tailed, we log-transformed these values.
To test the first prediction, we constructed a linear mixed-effects regression predicting trial-level speaker utterance length.
We included a fixed effect of repetition block within partner (1, 2, 3, 4), along with random intercepts and slopes for each participant and each tangram. 
We found that speakers reduced utterance length significantly over successive interactions with each individual partner, $b = -0.19,~t(34) = -9.88,~p < 0.001$.

To test the extent to which speakers revert to longer utterances at partner boundaries, we constructed another regression model.
We coded the repetition blocks immediately before and after each partner swap, and included it as a categorical fixed effect.
Because partner roles were randomized for each game, the same participant did not always serve as listener in both blocks, so in addition to tangram-level intercepts, we included random slopes and intercepts at the \emph{network} level (instead of the participant level).
As predicted, we found that utterance length increased significantly at the two partner swaps, $b = 0.43,~t(22) = 4.4,~p < 0.001$.

Finally, to test whether efficiency improves for the \emph{very first} interaction with each new partner, before observing any partner-specific information, we examined the simple effect of partner number at the trials immediately after the partner swap (i.e. $t=\{1,5,9\}$).
We found that participants gradually decreased the length of their initial descriptions with each new partner in their network, $b = -0.2,~t(516.5) = -6.07,~p < 0.001$ (see Fig. \ref{fig:results}A, final column), suggesting that speakers are bringing increasingly well-calibrated expectations into interactions with novel neighbors.
The partial-pooling model is the only model predicting all three of these effects.

\paragraph{Network convergence}

Now, we examine the \emph{content} of conventions and evaluate the extent to which alignment increased across the network over the three partner swaps. 
Specifically, we extend the same measure of alignment used in our simulations to natural language data by examining whether the intersection of words produced by different speakers was non-empty.
We excluded a list of common stop words (e.g. ``the'', ``both'') to focus on core conceptual content.
While this pure overlap measure provides a relatively weak notion of similarity, a more continuous measure based on the \emph{size} of the intersection or the string edit distance yielded similar results.

As in our simulation, the main comparison of interest was between currently interacting participants and participants who are not interacting: the partial-pooling model predicted that within-pair alignment should stay consistently high while (tacit) alignment between non-interacting pairs will increase. 
To test this prediction, we constructed a mixed-effects logistic regression including fixed effects of pair type (within vs. across), partner number, and their interaction.
We included random intercepts at the tangram level and maximal random effects at the network level (i.e. intercept, both main effects, and the interaction).
As predicted, we found a significant interaction ($b = -0.85, z = -5.69, p < 0.001$; see Fig. \ref{fig:results}B, final column).
Although different pairs in a network may initially use different labels, these labels begin to align over subsequent interactions. 

This finding is consistent with the primary prediction of interest for both the complete-pooling and partial-pooling model. 
These two models only pull apart for a secondary prediction concerning the transition from the first to second partner.
The complete-pooling model predicts a significant drop in \emph{within-pair} convergence from the first to second partner, due to the continued influence of the first partner, while the partial-pooling model predicts no drop. 
We found no evidence of such a drop in the empirical data ($z=-0.66, p = 0.511$), providing further evidence in favor of the full partial-pooling structure.

\subsection{Discussion}

Drawing on general principles of hierarchical Bayesian inference, CHAI suggests that conventions represent the shared structure that agents ``abstract away'' from partner-specific learning.
In this section, we evaluated the extent to which CHAI captured human generalization behavior in a natural-language communication experiment on small networks.
Unlike complete-pooling accounts, it allows for partner-specific common ground to override community-wide expectations given sufficient experience with a partner, or in the absence of strong conventions.
Unlike no-pooling accounts, it results in networks that are able to converge on shared conventions.

Partner-specificity and generalization presents an even steeper challenge for previous accounts than \textbf{P1}. 
It is not straightforward for previous interactive alignment or reinforcement learning accounts to explain patterns across partner boundaries without being augmented with additional social information.
If a particular semantic representation has been primed due to precedent in the preceding dialogue, then the shifting identity of the speaker should not necessarily alter its influence \cite{brennan2009partner,ferreira2012priming,ostrand2019repeat}. 
More sophisticated hierarchical memory retrieval accounts that represent different partners as different \emph{contexts} \cite<e.g>{polyn2009context,BrownSchmidtEtAl15_Contexts} may allow priming to be modulated in a partner-specific way, but such an account would presuppose that social information like partner identity is already a salient and relevant feature of the communicative environment.
Indeed, an  account assuming socially-aware context reinstatement for partner-specific episodic memories, and slower consolidation of shared features into population-level expectations, may be one possible process-level candidate for realizing our hierarchical computational-level model.

A frequent concern in prior work using repeated reference games is that improvements in communication over time are due to generic effects of task familiarity and repetition rather than interactive adaptation to a partner's language use \cite{HupetChantraine92_CollaborationOrRepitition}. As they get more practice with the task, speakers may simply get better overall at describing images and listeners may learn how to better identify target images. The effects we observe at partner boundaries show that something is being learned beyond pure familiarity with the task: if speakers and listeners were just learning to better describe and identify targets regardless of who their partner is, we would not expect these reversions. These partner-specificity effects clearly rule out the complete pooling model, but cannot rule out a no-pooling model combined with a practice effect. Under this  alternative possibility, partner-specific adaptation would be genuine, but the general decrease in utterance length and increase in accuracy with new partners would be due to practice rather than inductive generalization.
Our best current evidence against this practice-based explanation lies in our network convergence results: networks as a whole converge to similar short descriptions across partners, and different networks converge to different descriptions, indicating some gradual degree of transfer across partners. 
Future work may further address these concerns by including filler trials or by manipulating the length of interaction with each partner.

Our account also predicts that similar inductive learning mechanisms would operate not only across different partners but across different contexts containing different referents. 
By holding the partner constant across different contexts, rather than holding the context constant across different partners, it would be possible to test the extent to which additional experience along one axis of generalization would affect generalization along the other axis. 
Finally, one subtler point, which we believe is a rich direction for future research, is how generalization may still depend on the speaker’s beliefs about \emph{how partners are sampled}, manifested in their inductive biases at the community-level (Eq.~\ref{eq:top_level}).
If they believe they are in a tight-knit community where different partners are experts with the domain and have likely interacted with one another before, they may generalize differently than if they believe their community has higher turnover and many novices, brand-new to the task \cite{IsaacsClark87_ReferencesExpertsNovices}.


\section{Phenomenon \#3:  Conventions are shaped by communicative context}

\begin{figure*}[t]
\begin{center}
{\includegraphics[scale=1.6]{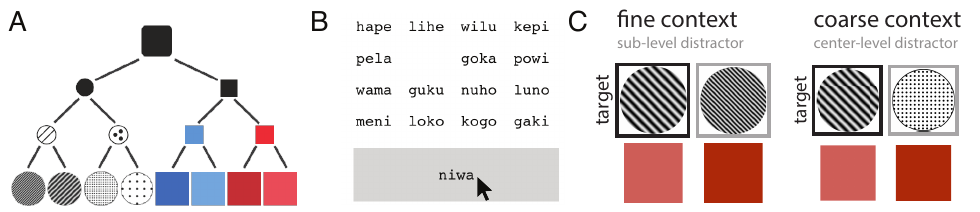}}
{\caption{{\emph{Context-sensitivity experiment.} (A) Targets are related to one another in a conceptual taxonomy. (B) Speakers choose between labels, where the label ``niwa'' has been selected. (C) Examples of \emph{fine} and \emph{coarse} contexts. In the \emph{fine} context, the target (marked in black) must be disambiguated from a distractor (marked in grey) at the same subordinate-level branch of the taxonomy.  In the \emph{coarse} context, the closest distractor belongs to a different branch of the center-level of the taxonomy (i.e. a spotted circle) such that disambiguation at the sub-ordinate level is not required. \label{fig:context_design}}}}
\end{center}
\end{figure*}

In the previous two sections, we evaluated a model of rapid, partner-specific learning that allows agents to form stable but arbitrary \emph{ad hoc} conventions with partners that gradually generalize to their entire community. 
The final phenomenon we consider is the way that \emph{ad hoc} conventions are shaped by the communicative needs of the context in which they form.
This phenomenon is most immediately motivated by recent findings that more informative or diagnostic words in the local referential context are significantly more likely to become conventionalized \cite{hawkins2020characterizing}.
For example, consider an initial description like ``the guy that looks like an ice skater with a leg up in front.''
A word like ``skater,'' which is distinctive of that single referent, is empirically more likely to persist in the resulting convention than words like ``guy'' or ``leg'' which are used in descriptions for multiple referents.
Our broader theoretical aim is to suggest that context-sensitivity in the \emph{synchronic} processes at play when individual dyads coordinate on  \emph{ad hoc} meanings may help to explain \emph{diachronic} balance of efficiency and expressivity in the long-term evolution of a community's lexicon, as highlighted by functionalist accounts like the Optimal Semantic Expressivity (OSE) hypothesis \cite{frankblogpost}. 

Briefly, when there is already a strong existing convention that is expected to be shared across the community, our model predicts that speakers will use it. 
New \emph{ad hoc} conventions arise precisely to fill \emph{gaps} in existing population-level conventions, to handle new situations where existing conventions are not sufficient to accurately and efficiently make the distinctions that are required in the current context. 
A corollary of this prediction is that \emph{ad hoc} conventions may only shift to expectations at the population level (and ultimately to population-level convergence) when those distinctions are consistently relevant across interactions with different partners\footnote{This follows by induction from the hierarchical generalization mechanisms evaluated for $\textbf{P2}$, which provide the pathway by which \emph{ad hoc} conventions become adopted by a larger community over longer time scales. Many \emph{ad hoc} conventions never generalize to the full language community simply because the contexts where they are needed are rare or variability across partners is too high. They must be re-negotiated with subsequent partners on an \emph{ad hoc} basis.}.
For example, while most English speakers have the general term  ``tree'' in their lexicon, along with a handful of subordinate-level words like ``maple'' or ``fir,'' we typically do not have conventionalized labels exclusively referring to each individual tree in our yards -- we are rarely required to refer to individual trees.
Meanwhile, we \emph{do} often have shared conventions (i.e. proper nouns) for individual people and places that a community regularly encounters and needs to distinguish among.
Indeed, this logic may explain why a handful of particularly notable trees do have conventionalized names, such as the Fortingall Yew, the Cedars of God, and General Sherman, the giant sequoia.

As a first step toward explaining these diachronic patterns in \emph{which} conventions form, we aim to establish in this section that our model allows a single dyad's \emph{ad hoc} conventions to be shaped by communicative context over short timescales.
Specifically, our model predicts that people will form conventions at the highest level of abstraction that is able to satisfy their communicative needs.
That is, when the local environment imposes a communicative need to refer to particular \emph{ad hoc} concepts (e.g. describing a particular tree that needs to be planted), communicative partners are able to coordinate on efficient lexical conventions for successfully doing so at the relevant level of abstraction (e.g. ``the mossy one'').

We begin by showing that this form of context-sensitivity naturally emerges from our model, as a downstream consequence of recursive pragmatic reasoning.
When a particular partner uses a label to refer to an object in a context, we can infer that they do not believe it ambiguously applies to distractors as well; otherwise, they would have known it would be confusing and chosen a different label.
We then empirically evaluate this prediction by manipulating which distinctions are relevant in an artificial-language repeated reference game building on \citeA{WintersKirbySmith14_LanguagesAdapt,winters2018contextual}, allowing us to observe the emergence of \emph{ad hoc} conventions from scratch.
In both the empirical data and our model simulations, we find that conventions come to reflect the distinctions that are functionally relevant for communicative success. 

\subsection{Model predictions: Simulation 3.1}

To evaluate the impact of context on convention formation, we require a different task than we used in the previous sections.
Those tasks, like most reference games in the literature on convention formation, used a discrete set of unrelated objects in a fixed context, $\{o_1, \dots, o_k\}$. 
In real referential contexts, however, targets are embedded in larger conceptual taxonomies, where some objects are more similar than others \cite{bruner1956study,collins1969retrieval,XuTenenbaum07_WordLearningBayesian}.
Here, we therefore consider a space of objects embedded in a three-level stimulus hierarchy with shape at the top-most level, color/texture at the intermediate levels, and frequency/intensity at the finest levels (see Fig.~\ref{fig:context_design}A). 
While we will use the full stimulus set in our empirical study, it is sufficient for our simulations to consider just one of the branches (i.e. just the four squares).
We populate the space of possible utterance meanings $P(\phi)$ with four meanings at the sub-ordinate level (one for each individual object, e.g. $\phi(u) =$ ``light blue square''), 2 meanings at the center-level (e.g. $\phi(u) =$ ``blue square''), 1 meaning at the super-ordinate level (e.g. $\phi(u) =$ ``square''). 
We allow for a ``null'' meaning with an empty extension to account for the possibility that some utterances are not needed, allowing the agent to effectively remove utterances from their vocabulary. 
We then populate the utterance space with 8 single-word labels (Fig.~\ref{fig:context_design}B).

Another important feature of real environments is that speakers do not have the advantage of a fixed context; the relevant distinctions change from moment to moment as different subsets of objects are in context at different times. 
This property poses a challenge for models of convention formation because the relevant distinctions cannot be determined from a single context; they must be abstracted over time.
We therefore only displayed two of the four possible objects on a given trial.
Distractors could differ from the target at various levels of the hierarchy, creating different types of contexts defined by the finest distinction that had to be drawn (e.g. Fig.~\ref{fig:context_design}C).  

Critically, we manipulated the prevalence of different kinds of contexts, controlling how often participants are required to make certain distinctions to succeed at the task. 
In the \emph{fine} condition, every context contained a subordinate distractor, requiring fine low-level distinctions to be drawn.
In the \emph{coarse} condition, contexts never contained subordinate distractors, only distractors that differed at the central level of the hierarchy (e.g. a blue square when the target is a red square).
For comparison, we also include a \emph{mixed} condition, where targets sometimes appear in \emph{fine} contexts with subordinate distractors and other times appear in \emph{coarse} contexts without them; the context type is randomized between these two possibilities on each trial.
We constructed the trial sequence identically for the three conditions. 
On each trial, we randomly sampled one of the four possible objects to be the target.
Then we sampled a distractor according to the constraints of the context type.
As before, the agents swapped roles after each trial. 
We ran 400 distinct trajectories with parameter settings of $\alpha_L=8, \alpha_S=8$ and memory discounting parameter of $\beta = 0.8$ (see Fig.~\ref{fig:partitionPrior} for results at other parameter values).

\begin{figure}[t]
\begin{center}
\includegraphics[scale=0.87]{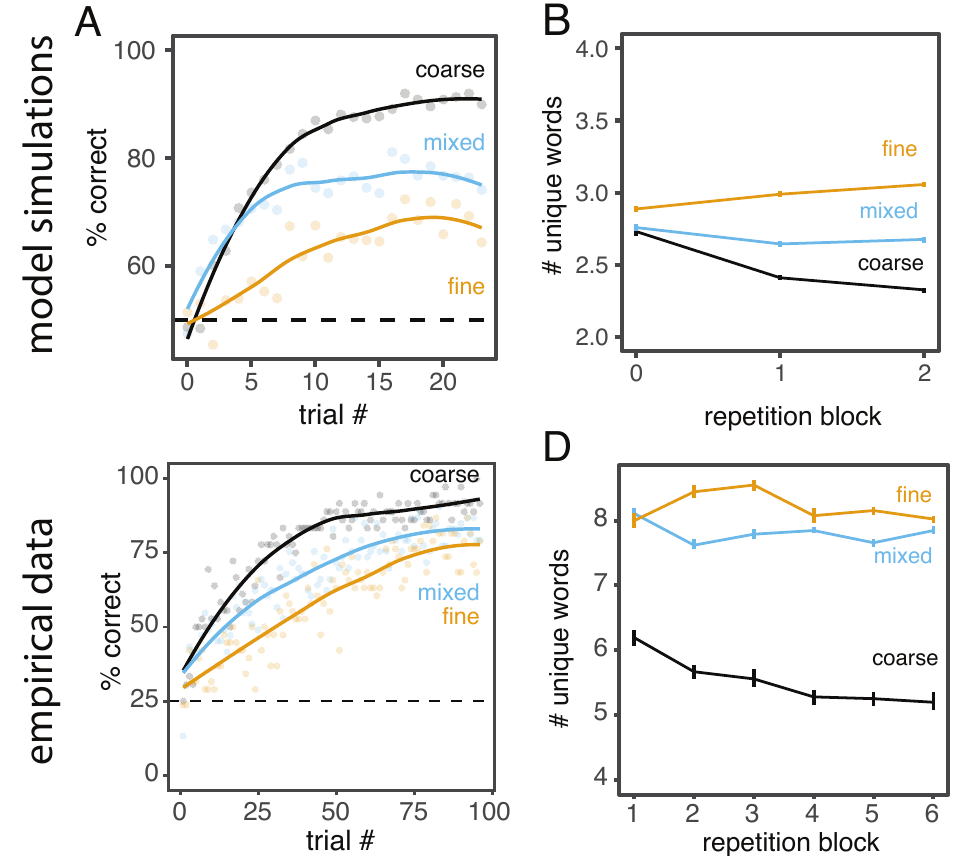}
\caption{\emph{Comparison of simulation results to empirical data}. (A) Agents in our simulation learn to coordinate on a successful communication system, but converge faster in the coarse condition than the fine condition.  (B) The number of unique words used by agents in each repetition block stayed roughly constant in the fine condition but decreased over time in the coarse condition. (C-D) The same metrics computed on our empirical data, qualitatively matching the patterns observed in the simulations. Each point is the mean proportion of correct responses by listeners; curves are nonparametric fits and error bars are bootstrapped 95\% CIs.}
\label{fig:sec2Results}
\end{center}
\end{figure}

\begin{figure}[h]
\begin{center}
\includegraphics[scale=0.9]{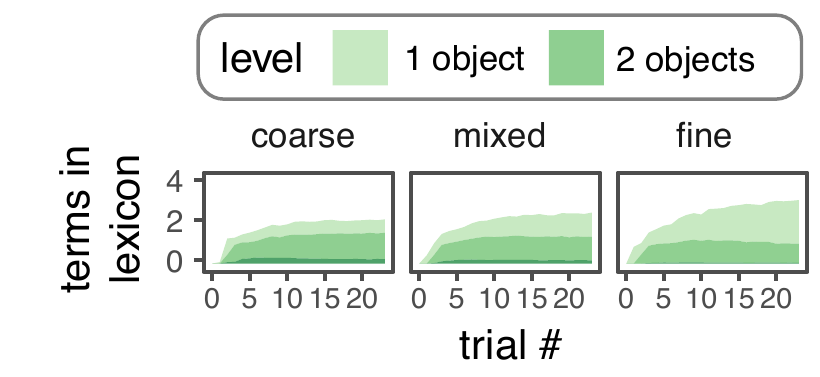}
\caption{\emph{Dynamics of lexical beliefs over time in model simulations.} Regions represent the average number of words at each level of generality in an agent's beliefs about the lexicon. Level of generality is determined by taking the MAP meaning. In the fine and mixed conditions, agents shift toward more subordinate terms.}
\label{fig:evolution}
\end{center}
\end{figure}

\subsubsection{Partners successfully learn to communicate}

First, we compare the model's learning curves across context conditions (Fig.~\ref{fig:sec2Results}A). 
In a mixed-effects logistic regression, we find that communicative accuracy steadily improves over time across all conditions, $b=0.72, z = 16.9, p<0.001$.
However, accuracy also differed across conditions: adding a main effect of condition significantly improves model fit, $\chi^2(2) = 9.6, p = 0.008$. 
Accuracy is significantly higher in the coarse condition than the fine condition $b=-0.71, z=9.3, p <0.001$ and marginally higher than the mixed condition.

\subsubsection{Lexical conventions are shaped by context}

As an initial marker of context sensitivity, we examine the effective vocabulary sizes used by speakers in each condition.
We operationalized this measure by counting the total number of unique words produced within each repetition block.
This measure takes a value of 8 when a different word is consistently used for every object, and a value of 1 when exactly the same word is used for every object.
In an mixed-effects regression model including intercepts and random effects of trial number for each simulated trajectory, we find an overall main effect of condition, with agents in the fine condition using significantly more words across all repetition blocks ($m = 4.7$ in \emph{coarse}, $m=6.5$ in \emph{fine,} $t = 4.5, p < 0.001$).
However, we also found a significant interaction: the effective vocabulary size gradually dropped over time in the coarse condition, while it stayed roughly constant in in the fine condition, $b = 0.18, t = 8.1, p < 0.001$, see Fig.~\ref{fig:sec2Results}B.

Next, we examine more closely the emergence of terms at different levels of generality.
We have access not only to the signaling behavior of our simulated agents, but also their internal \emph{beliefs} about their partner's lexicon, which allows us to directly examine the evolution of these beliefs from the beginning of the interaction.
At each time point in each game, we take the single meaning with highest probability for each word.
In Fig.~\ref{fig:evolution}, we show the proportion of words with meanings at each level of generality, collapsing across all games in each condition.
Qualitatively, we observe that agents begin by assuming null meanings (i.e. with an effectively empty vocabulary) but quickly begin assigning meanings to words based on their partner's usage.
In both conditions, basic-level meanings and subordinate-meanings are equally consistent with the initial data, but the simplicity prior prefers smaller effective vocabulary sizes.
After the first repetition block, however, agents in the coarse condition begin pruning out some of the subordinate-level terms and become increasingly confident of basic-level meanings.
Meanwhile, agents in the fine condition begin to disaggregate these basic-level terms into a greater number of subordinate-level meanings.

By the final trial, the proportion of basic-level vs.~subordinate-level terms is significantly different across the coarse and fine conditions.
Only 9\% of words had subordinate-level meanings (green) in the coarse condition, compared with 79\% in the fine condition, $\chi^2(1) = 436, p < 0.001$.
At the same time, 45\% of words had basic-level meanings (blue) in the coarse condition, compared with only 8\% in the fine condition, $\chi^2(1) = 136, p < 0.001$.
The remaining words in each condition were assigned the `null` meaning (red), consistent with an overall smaller effective vocabulary size in the coarse condition.
The diverging conventions across contexts are driven by Gricean expectations: because the speaker is assumed to be informative, only lexicons distinguishing between subordinate level objects can explain the speaker's behavior in the \emph{fine} condition.

\subsection{Experimental methods}

In this section, we evaluate our model's qualitative predictions about the effect of context on convention formation using an interactive behavioral experiment closely matched to our simulations.
We use a between-subjects design where pairs of participants are assigned to different communicative contexts and test the extent to which they converge on meaningfully different conventions.

\subsubsection{Participants}

We recruited 278 participants from Amazon Mechanical Turk to play an interactive, multi-player game. Pairs were randomly assigned to one of three different conditions, yielding $n=36$ dyads in the \emph{coarse} condition, $n=38$ in the \emph{fine} condition, and $n=53$ in the \emph{mixed} condition after excluding participants who disconnected before completion\footnote{This experiment was pre-registered at \url{https://osf.io/2hkjc/} with a target sample size of roughly 40 games per condition. We planned to include all participants for our accuracy analyses but then exclude participants who were still below 75\% accuracy on the final quarter of the task $(n=29$ pairs) for our analyses of the lexicon, to ensure post-test measurements could be interpreted as ``converging'' lexicons (as opposed to pairs who had lost interest or given up). We were later concerned that this exclusion could lead to spurious differences because convergence rates differed across conditions, but no results substantially changed depending on the exclusion criteria. All statistical tests in mixed-effects models reported in this section use degrees of freedom based on the Satterthwaite approximation \cite{luke2017evaluating}.}.

\subsubsection{Procedure \& Stimuli}
Participants were paired over the web and placed in a shared environment containing an array of four objects at a time (Fig.~\ref{fig:context_design}A) and a `chatbox' to choose utterances from a fixed vocabulary by clicking-and-dragging (Fig.~\ref{fig:context_design}B). On each trial, one player (the `speaker') was privately shown a highlighted target object and allowed to send a single word to communicate the identity of this object to their partner (the `listener'), who subsequently made a selection from the array. Players were given full feedback, swapped roles each trial, and both received bonus payment for each correct response.

We randomly generated distinct arrays of 16 utterances for each pair of participants (more than our model, which was restricted by computational complexity).
These utterances were created by stringing together consonant-vowel pairs into pronounceable 2-syllable words to reduce the cognitive load of remembering previous labels (see Fig.~\ref{fig:context_design}B).
These arrays were held constant across trials.
However, as in our simulations, the set of referents on each trial was manipulated in a between-subjects design to test the context-sensitivity of the resulting conventions. 
The trial sequence consisted of 6 blocks of 16 trials, for a total of 96 trials.
Each of the eight possible objects shown in Fig.~\ref{fig:context_design}A appeared as the target exactly twice per block, and was prevented from being shown twice in a row.
In addition to behavioral responses collected over the course of the game, we designed a post-test to explicitly probe players' final lexica. 
For all sixteen words, we asked players to select all objects that a word can refer to (if any), and for each object, we asked players to select all words that can refer to it (if any). 
This bidirectional measure allowed us to check the internal validity of the lexica reported checking mismatches between the two directions of the lexicon question (e.g.\ if they clicked the word `mawa' when we showed them one of the blue squares, but failed to click that same blue square when we showed `mawa').
We conservatively take a participant's final lexicon to be the \emph{intersection} of their word-to-object and object-to-word responses.


\begin{figure}[t]
\begin{center}
\includegraphics[scale=1]{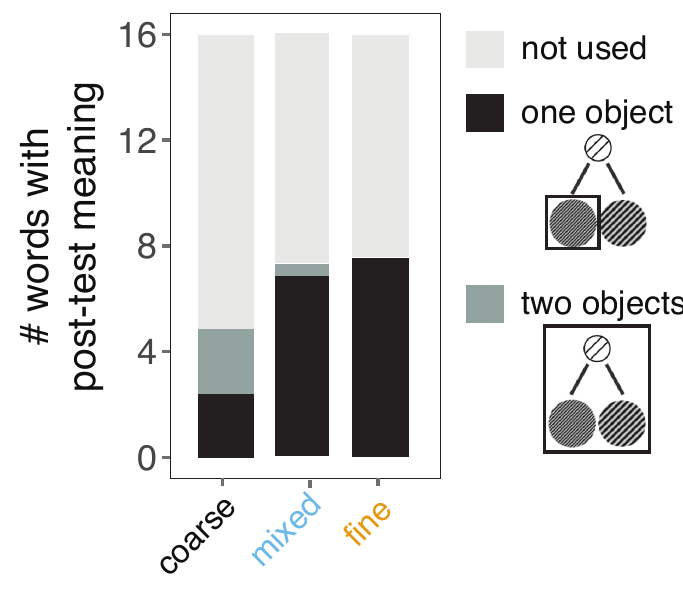}
\caption{\emph{Different lexicons emerge in different contexts.} Mean number of words, out of a word bank of 16 words, that human participants reported giving more specific meanings (black; applying to 1 object) or less specific meanings (dark grey; applying to 2 objects) in the post-test.}
\label{fig:sec2postTest}
\end{center}
\end{figure}

\subsection{Behavioral results}

\subsubsection{Partners successfully learn to communicate}

Although participants in all conditions began with no common basis for label meanings, performing near chance on the first trial (proportion correct $= 0.19$, 95\%~CI~$=~[0.13, 0.27]$), most pairs were nonetheless able to coordinate on a successful communication system over repeated interaction (see Fig.\ \ref{fig:sec2Results}C). 
A mixed-effects logistic regression on listener responses with trial number as a fixed effect, and including by-pair random slopes and intercepts, showed a significant improvement in accuracy overall, $z = 14.4, p < 0.001$. 
Accuracy also differed significantly \emph{across} conditions: adding an additional main effect of condition to our logistic model provided a significantly better fit, $\chi^2(2) = 10.8, p = 0.004$. 
Qualitatively, the \emph{coarse} condition was easiest for participants, the \emph{fine} condition was hardest, and the \emph{mixed} condition was in between.
These effects track the most important qualitative feature of our simulations -- our artificial agents were also able to successfully coordinate in both conditions, and did so more easily in the coarse condition than the fine condition. 
However, we found that the speed of coordination in the mixed and fine conditions was slower than predicted in our simulations.
The additional difficulty participants' experienced in the \emph{fine} condition may be due to additional motivational constraints, memory constraints, or other factors not captured in our model.

\subsubsection{Contextual pressures shape the lexicon}

We predicted that in contexts regularly requiring speakers to make fine distinctions among objects at subordinate levels of the hierarchy, we would find lexicalization of specific terms for each object (indeed, a one-to-one mapping may be the most obvious solution in a task with only 8 objects). 
Conversely, when no such distinctions were required, we expected participants to adaptively conventionalize more general terms that could be reused across different contexts.
One coarse signature of this prediction lies in the \emph{compression} of the resulting lexicon: less specific conventions should allow participants to achieve the same communicate accuracy with a smaller vocabulary.
We calculated the same measure of effective vocabulary size that we used in our simulations (Fig.~\ref{fig:sec2Results}D): the number of unique words produced in each repetition block. 
We then constructed a mixed-effects regression model predicting effective vocabulary size, including fixed effects of condition and six repetition blocks, with random intercepts and repetition block slopes for each dyad. 
First, we found an overall main effect of condition, with significantly fewer words used in the coarse condition $(m = 5.5)$ than the mixed ($m=7.9$, $t(95)=13.2$, $p <0.001$) or fine ($m=8.3$, $t(95) =13.3$, $p < 0.001$) conditions.
Consistent with our simulations, we also found a significant interaction between block and condition, with the coarse condition decreasing more over time than mixed ($b = 0.39$, $t(95) = 3.4$, $p < 0.001$) or fine ($b=0.36$, $t(95)=2.6$, $p=0.009$; see Fig.~\ref{fig:sec2Results}D).

What allowed participants in the coarse condition get away with fewer words in their lexicon while maintaining high accuracy?
We hypothesized that each word had a larger extension size. 
To test this hypothesis, we turned to our post-test survey.
We counted the numbers of `specific' terms (e.g.~words that refer to only one object) and `general' terms (e.g.~words that refer to two objects) in the post-test. 
We found that the likelihood of lexicalizing more general terms differed systematically across conditions.
Participants in the coarse condition reported significantly more general terms ($m=2.3$) than in the mixed ($m=0.47$, $t(91.8)= 8.8$, $p < 0.001$) or fine ($m = 0.04$, $t(90.2) = 9.2$, $p < 0.001$) conditions, where lexicons contained almost exclusively specific terms.
Using the raw extension size of each word as the dependent variable instead of counts yielded similar results.
Indeed, the modal system in the fine condition was exactly eight specific terms with no more general terms, and the modal system in the coarse condition was exactly four general terms (red, blue, striped, spotted) with no specific terms.
However, many individual participants reported a mixture of terms at different levels of generality (see Appendix Fig.~\ref{fig:mixtureOfTerms}). 



%

\subsection{Discussion}

There is abundant evidence that languages adapt to the needs of their users.
Our model provides a cognitive account of how people coordinate on \emph{ad hoc} linguistic conventions that are adapted to their immediate needs.
In this section, we evaluated predictions about these context effects using new data from a real-time communication task.
When combined with the generalization mechanisms explored in the previous section, such rapid learning within dyadic interactions may be a powerful contributor allowing languages to adapt at the population-level over longer time scales.

Previous studies of convention formation have addressed context-sensitivity in different ways.
In some common settings, there is no explicit representation of context at all, as in the task known as the ``Naming Game'' where agents coordinate on names for objects in isolation \cite{steels2012experiments,baronchelli2008depth}. 
In other settings, communication is situated in a referential context, but this context is held constant, as in Lewis signaling games \cite{lewis_convention:_1969} where agents must distinguish between a fixed set of world states \cite{skyrms2010signals,BrunerEtAl14_LewisConventions}.
Finally, in the more sophisticated Discrimination Game \cite{steels2005coordinating,baronchelli2010modeling}, contexts were randomly generated on each trial, but have not been manipulated to assess context-sensitivity of the convention formation process.

%
In other words, to the extent that context-sensitivity has been addressed by existing models, it has been implicit. 
Models using simple update rules have accounted for local referential context with a \emph{lateral inhibition} heuristic used by both the speaker and listener agents \cite{franke2012bidirectional,steels2005coordinating}.
If communication is successful, the connection strength between the label and object is not only increased, the connection between the label and competing objects (and, similarly, between the object and competing labels) is explicitly \emph{decreased} by a corresponding amount.
This lateral inhibition heuristic is functionally similar to our pragmatic reasoning mechanism, in terms of allowing the agent to learn from negative evidence (i.e. the speaker's choice \emph{not} to use a word, or the listener's choice \emph{not} to pick an object). 
Under our inferential framework, however, this form of statistical preemption emerges as a natural consequence of normative Gricean principles of pragmatic reasoning rather than as a heuristic (see also Appendix C for similar results using a alternative priors.)

\section{General Discussion}

Communication in a variable and non-stationary landscape of meaning creates unique computational challenges.
To address these challenges, we advanced a hierarchical Bayesian approach  in which agents continually adapt their beliefs about the form-meaning mapping used by each partner, in turn.
We formalized this approach by integrating three core cognitive capacities in a probabilistic framework: representing initial uncertainty about what a partner thinks words mean \textbf{(C1)}, partner-specific adaptation based on observations of language use in context \textbf{(C2)}, and hierarchical structure for graded generalization to new partners \textbf{(C3)}.
This unified model resolves several puzzles that have posed challenges for prior models of coordination and convention formation: why referring expressions shorten over repeated interactions with the same partner \textbf{(P1)}, how partner-specific common ground coexists with the emergence of conventions at the population level \textbf{(P2)}, and how context shapes which conventions emerge \textbf{(P3)}.

We conclude by raising three broader questions that arise from the perspective of our model, each suggesting pathways for future work: (1) to what extent is \emph{ad hoc} convention formation in adults the same as word learning in children and how is it different? (2) to what extent do the proposed mechanisms depend on the communication modality? and (3) which representations are involved in adaptation at a process-level? 

\subsection{Continuity of language learning across development}

CHAI aims to shift the central computational problem of communication from transmission to learning and adapation. 
Although it is intended as a theory of \emph{adult} communication among mature language users, our emphasis on learning has much in common with theories of language acquisition in development.
Could the basic cognitive mechanisms allowing adults to coordinate on conventions be the same as those supporting learning in children?
In other words, is it possible that adults never stop learning language and simply develop better-calibrated priors? 
In this section, we discuss three possible implications of viewing language acquisition in terms of social coordination and convention formation, which may help to further unify models of adult communication with those of language learning \cite<e.g.>{XuTenenbaum07_WordLearningBayesian,FrankGoodmanTenenbaum09_Wurwur,bohn2019pervasive}.

First, developmental paradigms have typically focused on variability and generalization across referential contexts (e.g. in cross-situational word learning) rather than variability and generalization across speakers  \cite{siskind1996computational,regier2005emergence,smith2014unrealized,yurovsky2015integrative}. 
Yet it is increasingly apparent that children are able to track \emph{who} produced the words they are learning and use this information when generalizing.
For example, bilingual children learn to expect different languages to be used by different speakers and even infants are sensitive to coarse social distinctions based on foreign vs. native language \cite{KinzlerDupouxSpelke07_LanguageGroups}, or accent \cite{KinzlerEtAl09_AccentRace}. 
Children are also sensitive to the reliability of individual speakers. 
For example, young children may limit the generalizability of observations from speakers who use language in idiosyncratic ways, such as a speaker who calls a ball a ``dog'' \cite{koenig2010sensitivity,luchkina2018eighteen}, and may even retrospectively update their beliefs about earlier evidence from a speaker after observing such idiosyncracies \cite{dautriche2021}. 
Such discounting of idiosyncratic speakers may be understood as an instance of the same inductive problem that convention formation poses for adults in \textbf{P2}.
Unlike complete-pooling models, which predict that all observations should be equally informative about a community's conventions, CHAI predicts that children should be able to explain away ``outliers'' without their community-level expectations being disrupted.
One novel prediction generated by our account is that children should be able to accommodate idiosyncratic language \emph{within} extended interaction with the same speaker (e.g. continue to pretend the ball is called ``dog,'' given partner-specific common ground) while also limiting generalization of that convention \emph{across} other speakers.

Second, CHAI emphasizes the importance of representing lexical \emph{uncertainty} (\textbf{C1}), capturing expected variability in the population beyond the point estimates assumed by traditional lexical representations. 
But how do children calibrate their lexical uncertainty?
The number of distinct speakers in a child's environment may play a key role, by analogy to the literature on talker variability \cite{creel2011talker,clopper2004effects}.
Exposure to fewer partners may result in weaker  or mis-calibrated priors \cite<e.g.>{lev2017talking}.
If an idiosyncratic construction is over-represented in the child's environment, they may later be surprised to find that it was specific to their household's lexicon and not shared by the broader community  \cite<see>[Chap. 6]{Clark09_FirstLanguageAcquisition}. 
Conversely, however, hierarchical inference predicts a blessing of abstraction \cite{GoodmanUllmanTenenbaum11_TheoryOfCausality}: under certain conditions, reliable community-level conventions may be inferred even with relatively  sparse observations from each partner.
To resolve these questions, future work will need to develop new methods for eliciting children's expectations about partner-specificity and variability of meanings.

Third, our work suggests a new explanation for why young children struggle to coordinate \emph{ad hoc} conventions with one another in repeated reference games \cite{GlucksbergKraussWeisberg66_DevoRefGames,KraussGlucksberg69_DevoReferenceGames,KraussGlucksberg77_SocialNonsocialSpeech,matthews2007toddlers}. 
Early explanations appealed to rigidity in the child's perspective that prevented adaptation.
Yet subsequent findings that children could not even interpret their own utterances after a delay \cite{asher1976children} suggest that the challenge may instead stem from production quality and the lack of coordinating 'signal'. 
Children may either be unable to anticipate how much information is required for their partner to discriminate the referent, or struggle to access those more complex formulations. 
In terms of our model, children's lexical priors may be weaker than adults': without existing conventions for describing the novel objects in their vocabulary, their utterances are dispersed widely over easier-to-access ``good-enough'' formulations \cite{goldberg2019explain}.
Indeed, when children are paired with their caregivers rather than peers, they easily coordinate on new conventions \cite{LeungEtAl20_Pacts}.
Adults helped to interactively scaffold the conventions, both by proactively seeking clarification when in the listener role \cite<e.g.>{anderson1994interactive} and by providing more descriptive labels when in the speaker role, which children immediately adopted\footnote{It may be observed that agents in our simulations were still able to quickly coordinate despite being initialized with weak priors, but they had the benefit of using feedback from the referential task, as well as small, shared vocabularies. In the paradigms used by \citeA{KraussGlucksberg69_DevoReferenceGames}, young children did not have access to such information and may have struggled to search their vocabulary for better candidates even if they did, especially under time pressure \cite<e.g.>{glucksberg1967people}. This kind of accessibility consideration has previously been instantiated in computational models via the cost term $c(u)$, but further work on convention formation in developmental samples may benefit from a more fine-grained, process model of production.}.
From this perspective, \emph{ad hoc} conventions may not be so different from other settings where children look to adults for guidance and rapidly adopt new conventions to talk about new things \cite<e.g.>{carey1978acquiring,heibeck1987word}. 


\subsection{The role of communication modality}

One of our core claims is that the basic learning mechanisms underlying coordination and convention formation are domain-general.
In other words, we predict that there is nothing inherently special about spoken or written language: any system that humans use to communicate should display similar \emph{ad hoc} convention formation dynamics because in every case people will be trying to infer the system of meaning being used by their partners. 
Directly comparing behavior in repeated reference games across different modalities is therefore necessary to determine which adaptation effects, if any, are robust and attributable to modality-general mechanisms.
In fact, there has been significant progress in understanding the dynamics of adaptation during communication in the  graphical modality \cite{GarrodFayLeeOberlanderMacLeod07_GraphicalSymbolSystems,TheisenEtAl10_SystematicityArbitrariness,hawkins2019disentangling}, the gestural modality \cite{FayListerEllisonGoldinMeadow13_GestureBeatsVocalization,motamedi2019evolving,bohn2019young} and other \emph{de novo} modalities \cite{Galantucci05_EmergenceOfCommunication,RobertsGalantucci12_DualityOfPatterning,RobertsEtAl15_IconocityOnCombinatoriality,VerhoefRobertsDingemanse15_Iconicity,VerhoefEtAl16_TemporalLanguage,kempe2019adults}.

CHAI views the similarities and differences between modalities through the lens of the hierarchical \emph{priors} we have built up across interactions with different individuals.
For example, in the verbal modality, the tangram shapes from \citeA{ClarkWilkesGibbs86_ReferringCollaborative} are highly ``innominate'' \cite<meaning empirically difficult to name;>{HupetEtAl91_CodabilityReference,zettersten2020finding} -- most people do not have much experience naming or describing them with words, so relevant priors are weak and local adaptation plays a greater role.
In the graphical modality, where communication takes place by drawing on a shared sketchpad, people can be expected to have a stronger prior rooted in assumptions about shared perceptual systems and visual similarity \cite{fan2018common}.
Drawing a quick sketch of the tangram's outline may suffice for understanding.
Other referents have precisely the opposite property: to distinguish between natural images of dogs, people may have strong existing conventions in the linguistic modality (e.g. `husky', `poodle', `pug') but making the necessarily fine-grained visual distinctions in the graphical modality may be initially very costly for novices \cite{fan2020pragmatic}, requiring the formation of local conventions to achieve understanding \cite{hawkins2019disentangling}. 
The gestural modality also has its own distinctive prior, which also allows communicators to use time and the space around them to convey mimetic or depictive meanings that may be difficult to encode verbally or graphically \cite{goldin-meadow_role_1999,clark2016depicting,mcneill1992hand}. 
We therefore suggest that differences in production and comprehension across modalities may be understood by coupling modality-specific priors with modality-generic learning mechanisms.

\subsection{Process-level mechanisms for adaptation}

Finally, while we have provided a computational-level account of coordination and convention formation in terms of hierarchical inference, there remain many possible process-level mechanisms that may perform this computation.
In this section, we discuss two interlocking process-level questions which emphasize current limitations and areas of future work: (1) exactly which representations should be adapted? and (2) what is required to scale models of adaptation to more naturalistic language? 

\subsubsection{Which representations are adapted?}

While our model formulation focused on adaptation at the level of lexical meaning (i.e. inferences about $\phi$, representing different possible lexical meanings), this is only one of many internal representations that may need to be adapted to achieve successful coordination. 
Three other possible representational bases have been explored in the literature. 

First, it is possible that adaptation takes place upstream of the lexicon, directly implicating perceptual or \emph{conceptual} representations \cite{GarrodAnderson87_SayingWhatYouMean,HealeySwobodaUmataKing07_GraphicalLanguageGames}
That is, there may be uncertainty about how a particular partner construes the referent itself, and communication may require constructing a shared, low-dimensional conceptual space where the relevant referents can be embedded \cite{stolk2016conceptual}.
This is particularly clear in the classic maze task \cite{GarrodAnderson87_SayingWhatYouMean} where giving effective spatial directions requires speakers to coordinate on what spatial representations to use (e.g. paths, coordinates, lines, or landmarks). 

Second, it is possible that adaptation takes place even further upstream, at the level of \emph{social} representations \cite{jaech2018low}.
Rather than directly updating beliefs about lexical or conceptual representations, we may update a holistic representation of the partner themselves (e.g. as a ``partner embedding'' in a low-dimensional vector space) that is used to retrieve downstream conceptual and lexical representations. 
Under this representational scheme, the mapping from the social representation to particular conventions is static, and \emph{ad hoc} adaptation is limited to learning where a particular partner belongs in the overall social space.

Third, expectations about other lower-level features may also be adapted through interaction, such as a partner's word frequencies \cite{louwerse2012behavior}, syntax \cite{gruberg2019syntactic,levelt1982surface}, body postures \cite{lakin2003using}, speech rate \cite{giles1991contexts}, or even informational complexity \cite{abney2014complexity}.
This level of adaptation may lead some forms to become more accessible or entrenched in memory over time, possibly allowing partner identity to be used as a retrieval cue (e.g. \citeNP{horton2005impact,horton2007influence,horton_revisiting_2016}; but see \citeNP{brown2014influence}).

\subsubsection{Computational tractability and scalability}

While a fully Bayesian formulation elegantly formalizes the computational-level inference problem at the core of the CHAI account, this formulation faces a number of limitations. 
For one, it is clearly intractable \cite{van2008tractable,van2019cognition}: the posterior update step in Eq.~\ref{eq:joint_inference} grows increasingly intensive as the space of possible utterances and meanings grows \cite{scalingupmodels}.
The intractability problem also raises a scalability problem: does CHAI provide any guidance toward building artificial agents that are actually able to adapt to human partners as humans do with one another?
Through this applied lens, a number of recent efforts have focused on developing algorithms for state-of-the-art neural networks that tractably scale to arbitrary natural language (e.g. referring expressions using the full vocabulary of an adult language user) and arbitrary visual input (e.g. sensory impressions of novel objects such as tangrams).

For example, building on recent formal connections between hierarchical Bayes and gradient-based meta-learning approaches in machine learning \cite{grant_recasting_2018}, the algorithm proposed by \citeA{hawkins2019continual} (1) relaxes the full community-level prior over $\Theta$ to a point estimate and (2) replaces the difficult integral in the posterior update with a fixed number of (regularized) gradient update steps.
Another recent proposal builds on connections to classical exemplar-based algorithms \cite{nosofsky1984choice}: an agent's lexical expectations at time $t$ may be determined via weighted similarity to memory traces of lexical items used by different partners in the past \cite{shi2010exemplar}, where similarity is computed by a neural network. 
While such algorithms cannot fix the intractability of the Bayesian formulation \cite{kwisthout2011bayesian}, and their precise correspondence to constraints on the computational-level theory remain unexplored, they nonetheless provide promising algorithmic instantiations of the CHAI account. 
When lexical meaning is represented by the parameters of a neural network, conventions can be interpreted as (meta-)learned \emph{initializations} used for new partners and coordination is partner-specific \emph{fine-tuning} or \emph{domain adaptation} of vector representations. 

Neural network instantiations also provide a possible pathway toward addressing the lack of incrementality in the fully Bayesian formulation. 
As more scalable implementations of pragmatic reasoning have proliferated in machine learning \cite{vogel2013emergence,AndreasKlein16_Pragmatics,monroe_colors_2017,shen2019pragmatically,takmaz2020refer} it has been natural to use incremental architectures  \cite{augurzky2019gricean,cohn2018pragmatically,cohn2019incremental,waldon2021modeling}. 
However, there remain a number of limitations to address in future work, including how to incorporate incremental \emph{feedback} into lexical updates (e.g. backchannel responses or interruptions), how to define a more satisfying notion of \emph{compositional semantics} for incrementally constructed utterances, and how to maintain representations of partner-specific parameters alongside community-wide parameters in memory.

\paragraph{Conclusion}

How do we manage to understand one another?
We have argued that successful communication depends not just on transmission but on continual learning across multiple timescales. 
We must coordinate on meaning through common ground with individual partners but also abstract these experiences away to represent stable \emph{conventions} and norms that generalize across our communities.
Like other socially-grounded knowledge, language is not a rigid dictionary that we acquire at an early age and deploy mechanically for the rest of our lives. 
Nor do languages only change over the slow time-scales of inter-generational drift.
Language is a means for communication -- a shared interface between minds -- and as new \emph{ad hoc} concepts arise, new \emph{ad hoc} conventions must be formed to solve the new coordination problems they pose.
In other words, we are constantly learning language. 
Not just one language, but a family of related languages, across interactions with each partner. 

\begin{quote}
\emph{Let us conclude not that ‘there is no such thing as a language’ that we bring to interaction with others. Say rather that there is no such thing as the one total language that we bring. We bring numerous only loosely connected languages from the loosely connected communities that we inhabit.} \cite{hacking1986nice}
\end{quote}

\section{\bf Acknowledgments}
\small

Thanks to Herb Clark, Judith Degen, Natalia V\'elez, Rosa Cao, Hyo Gweon, Judith Fan, Dan Yamins, Chris Potts, Iris van Rooij, Mark Blokpoel, Marten van Schijndel, and Josh Armstrong for helpful discussions.
This work was supported by NSF grant \#1911835 to RDH, AEG, and TDG.

\bibliography{ref}
\bibliographystyle{apacite-no-initials}

\renewcommand{\thefigure}{A\arabic{figure}}
\renewcommand{\thetable}{A\arabic{table}}
\setcounter{table}{0}
\setcounter{figure}{0}

\begin{table*}[th!]
\centering
\resizebox{\textwidth}{!}{%
\begin{tabular}{@{}lll@{}}
\toprule
\textbf{}                                   & \textbf{Parameter}                                              & \multicolumn{1}{l}{\textbf{Example parameter settings}}                 \\ \midrule
\multirow{11}{*}{\textbf{Partner design}}    & \multirow{3}{*}{What feedback is provided?}                & - no feedback at all                                                  \\
                                            &                                                                 & - only correct/incorrect                                                   \\
                                            &                                                                 & - real-time responses from partner                                         \\ \cmidrule(l){2-3} 
                                            & \multirow{3}{*}{Are you playing with the same partner?}         & - same partner for whole game                                              \\
                                            &                                                                 & - swap out partners every round                                            \\
                                            &                                                                 & - swap after $k$ rounds                                                    \\ \cmidrule(l){2-3} 
                                            & \multirow{3}{*}{What do you know about your partner?}         & - anonymous stranger                                              \\
                                            &                                                                 & - stranger with perceptual information                                            \\
                                            &                                                                 & - close friend                                                    \\ \cmidrule(l){2-3}                                             
                                            & \multirow{2}{*}{How consistent are roles across repetitions?}   & - consistent director/matcher                                              \\
                                            &                                                                 & - alternate roles each round                                               \\ \midrule
\multirow{7}{*}{\textbf{Stimulus design}}   & \multirow{2}{*}{How familiar are targets?}                      & - very familiar: colors, household objects                                 \\
                                            &                                                                 & - not at all familiar: tangrams, novel line drawings                       \\ \cmidrule(l){2-3} 
                                            & \multirow{2}{*}{How complex are targets?}                       & - very complex: busy visual scenes, clips of music                         \\
                                            &                                                                 & - not at all complex: geometric drawings                                   \\ \cmidrule(l){2-3} 
                                            & \multirow{3}{*}{How consistent are targets across repetitions?} & - exact same image of object                                               \\
                                            &                                                                 & - different pose/view of same object                                       \\
                                            &                                                                 & - different objects from same neighborhood                                 \\ \midrule
\multirow{5}{*}{\textbf{Context design}}    & \multirow{2}{*}{How similar are distractors to the target?}     & - very similar: same basic-level category                                  \\
                                            &                                                                 & - not at all similar: other categories                                     \\ \cmidrule(l){2-3} 
                                            & What is the size of context?                                    & - between 2 and 21                                                         \\ \cmidrule(l){2-3} 
                                            & \multirow{2}{*}{How consistent is context across repetitions?}  & - exact same context each round                                            \\
                                            &                                                                 & - randomized context (sometimes far, sometimes close)                      \\ \midrule
\multirow{3}{*}{\textbf{Repetition design}} & How many repetitions per target?                                & - between 3 and 100                                                        \\ \cmidrule(l){2-3} 
                                            & \multirow{2}{*}{What is spacing between repetitions?}           & - block structure                                                          \\
                                            &                                                                 & - sequential structure with interspersed contexts                          \\ \midrule
\textbf{Modality design}                    & What medium is used for communication?                          & \begin{tabular}[c]{@{}l@{}}- text\\ - audio\\ - gesture\\ - drawing\end{tabular} \\ \bottomrule
\end{tabular}%
}
\caption{\normalfont{Proposed parameterization for repeated reference games, each of which theoretically impacts the formation of conventions.}}
\label{table:parameters}
\end{table*}

\normalsize

\section*{Appendix A: Details of RSA model}

Our setting poses several technical challenges for the Rational Speech Act (RSA) framework.
In this Appendix, we describe these challenges in more detail and justify our choices.

\subsection{Action-oriented vs. belief-oriented listeners}

First, both agents are ``action-oriented,'' in the sense that they behave proportional to the utility of different actions, according to a soft-max normalization $\sigma(U(z)) = e^{U(z)}/\sum e^{U(z)}$.
This contrasts with some RSA applications, where the listener is instead assumed to be ``belief-oriented,'' simply inferring the speaker's intended meaning without producing any action of their own \cite{qing2015variations}.

\subsection{Placement of uncertainty}

Second, our instantiation of lexical uncertainty differs subtly from the one used by \citeA{bergen_pragmatic_2016}, which placed the integral over lexical uncertainty at a \emph{single} level of recursion (specifically, within a pragmatic listener agent).
Instead, we argue that it is more natural in an interactive, multi-agent setting for each agent to maintain uncertainty at the highest level, such that each agent is reasoning about their \emph{partner}'s lexicon regardless of what role they are currently playing.

\subsection{Handling degenerate lexicons}

Finally, when we allow the full space of possible lexicons $\phi$, we must confront degenerate lexicons where an utterance $u$ is literally false of every object in context, i.e. where $\mathcal{L}_\phi(o, u) = 0$ for all $o\in \mathcal{C}$. 
In this case, the normalizing constant in Eq.~\ref{eq:RSA} is zero, and the literal listener distribution is not well-defined.
A similar problem may arise for the $S_1$ distribution.

Several solutions to this problem were outlined by \citeA{bergen_pragmatic_2016}.
One of these solutions is to use a `softer' semantics in the literal listener, where a Boolean value of false does not strictly rule out an object but instead assigns a very low numerical score, e.g. 
$$\mathcal{L}_\phi(o,u) = \left\{ \begin{array} {rl} 1 & \textrm{if }o \in \phi(u) \\ \epsilon & \textrm{o.w.} \end{array}\right.$$
Whenever there is at least one $o\in\mathcal{C}$ where $u$ is true, this formulation will assign negligible listener probability to objects where $u$ is false, but ensures that the normalization constant is non-zero (and specifically, that the distribution is uniform) when $u$ is false for all objects.

While this solution suffices for one-shot pragmatics under lexical uncertainty, where $\epsilon$ may be calibrated to be appropriately large, it runs into several technical complications in an iterated setting.
First, due to numerical overflow at later iterations for some parameter values, elements may drop entirely out of the support at higher levels of recursion (e.g. $L_1$), leading the normalization constant to return to zero. 
Second, this `soft' semantics creates unexpected and unintuitive consequences at the level of the pragmatic speaker. 
After renormalization in $L_0$, an utterance $u$ that fails to refer to any object in context is also by definition equally \emph{successful} for all objects (i.e. evaluating to $\epsilon$ for every object), leading to a uniform selection distribution.
However, this assumption has the unintuitive consequence that $S_1$'s utility of using an utterance known to be false of the target may be the same as an utterance known to be true.

Instead of injecting $\epsilon$ into the lexical meaning, we ensure that the normalization constant is well-defined by adapting another method suggested by \citeA{bergen_pragmatic_2016}.
First, we add a `null' object to every context so that, even if a particular utterance is false of every real object in context, it will still apply to the null object, assigning the true target a negligible probability of being chosen.
Intuitively, this null object can be interpreted as recognizing that the referring expression has a referent but it is not in context, i.e. a `failure to refer,' and effectively prevents $L_0$ from assigning belief to a referent for which the utterance is literally false.
Note that this case is distinct from the case of a \emph{contradiction}, which arises when defining the meaning of multi-word utterances in Section \textbf{P1}.

Second, we add an explicit noise model at every level of recursion.
That is, we assume every agent has a probability $\epsilon$ of choosing a random element of their support, ensuring a fixed non-zero floor on the likelihood of each element that is constant across levels of recursion.
Formally this corresponds to a mixture distribution, e.g. 
$$L_0^{\epsilon}(o|u,\phi) = \epsilon \cdot P_{unif}(o) + (1-\epsilon) \cdot L_0(o|u,\phi)$$
$$S_1^{\epsilon}(u|o,\phi) = \epsilon \cdot P_{unif}(u) + (1-\epsilon) \cdot S_1(u|o,\phi)$$

\subsection{Marginalizing over $\phi_k$}
Another theoretical question arises about exactly how speaker and listener agents ought to marginalize over their uncertainty about $\phi_k$ when selecting actions (Eq.~\ref{eq:marginalized}). 
In our formulation, the expectation is naturally taken over the entire \emph{utility} each agent is using to act, i.e. if the speaker and listener utilities are defined to be 
$$
\begin{array}{rcl}
U_L(o;u, \phi_k) & = &  \log S_1(u|o, \phi_k)\\
U_S(u;o, \phi_k) & = &  (1-w_C)\log L_0(o|u, \phi_k) - w_C \cdot c(u) \\
\end{array}
$$
then the expectation is taken as follow:
$$
\begin{array}{rcl}
L(o|u) & \propto & \exp\left\{w_L\int P_L(\phi_k |D_k)\cdot U_L(u; o, \phi_k) \, d \phi_k\right\}\\
S(u|o) & \propto & \exp\left\{w_S\int P_S(\phi_k | D_k) \cdot U_S(u; o, \phi_k) \, d \phi_k\right\} \\
\end{array}
$$
This formulation may be interpreted as each agent choosing an action proportional to its expected utility across different possible values of $\phi_k$, weighted by the agent's current posterior beliefs about the lexicon their partner is using.

This formulation contrasts with the one suggested by \citeA{bergen_pragmatic_2016}, which assumes the expectation takes places at a single level of recursion, say the $L_1$, as above, and then derives the other agent's behavior by having them reason directly about this marginalized distribution, e.g.
$$
\begin{array}{rcl}
U_{alt1}(u;o) & = & (1-w_C) \cdot \log L(o|u) - w_C \cdot c(u)\\
S_{alt1}(u|o) & \propto & \exp\left\{w_S \cdot U_{alt1}(u;o)\right\} \\
\end{array}
$$
where $L(o|u)$ is defined as above.
This formulation may be interpreted as an assumption on the part of the speaker that the listener is already accounting for their own uncertainty, and best responding to such a listener.
Isolating lexical uncertainty over $\phi$ to a single level of recursion is a natural formulation for one-shot pragmatic phenomena, where additional layers of recursion can build on top of this marginal distribution to derive implicatures.
However, the interpretation is messier for the multi-agent setting, since it (1) induces an asymmetry where one agent considers the other's uncertainty but not vice versa, and (2) requires the speaker to use their own current posterior beliefs to reason about the listener's marginalization.

A third possible variant is to place the expectation outside the listener distribution but inside the speaker's informativity term, i.e..
$$
\begin{array}{rcl}
L_{avg} & = & \int P(\phi_k | D_k)\cdot L_0(o|u, \phi_k) d\phi_k \\
U_{alt2}(u;o) & = & (1-w_C) \cdot \log L_{avg}(o|u) - w_C \cdot c(u)\\
S_{alt2}(u|o) & \propto & \exp\left\{w_S \cdot U_{alt2}(u;o)\right\} \\
\end{array}
$$
The interpretation here is that the speaker first derives a distribution representing how a listener would respond \emph{on expectation} and then computes their surprisal relative to this composite listener.
While this variant is in principle able to derive the desired phenomena, it can be shown that it induces an unintuitive initial bias under a uniform lexical prior, since the logarithm cannot distribute over the integral in the normalization constant. 
This bias is most apparent in the case of context-sensitivity (Simulation 3).

Mathematically, the difference between these alternatives is whether the speaker's uncertainty about $\phi_k$ goes inside the renormalization of $L(o|u)$ (as in $S_{alt1}$), outside the renormalization but inside the logarithm (as in $S_{alt2}$), or over the entire utility (as in our chosen formulation).
While other formulations are conceivable, we argue that marginalizing over the entire utility is not only the most natural but also normatively correct under Bayesian decision theory. 
When an agent is uncertain about some aspect of the decision problem, rational choice requires the agent to optimize expected utility marginalizing over subjective uncertainty, as in our formulation. 

\subsection{Inference details} 

We have implemented our simulations in the probabilistic programming language WebPPL \cite{GoodmanStuhlmuller14_DIPPL}.
All of our simulations iterate the following trial-level loop: (1) sample an utterance from the speaker's distribution, given the target object, (2) sample an object from the listener's object distribution, given the utterance produced by the speaker, (3) append the results to the list of observations, and (4) update both agents' posteriors, conditioning on these observations before continuing to the next trial.
To obtain the speaker and listener distributions (steps 1-2; Eq.~\ref{eq:RSA}), we always use exhaustive enumeration for exact inference.
We would prefer to use enumeration to obtain posteriors over lexical meanings as well (step 4; Eq.~\ref{eq:joint_inference}), but as the space of possible lexicons $\phi$ grows, enumeration becomes intractable.
For simulations related to \textbf{P2} and \textbf{P3}, we therefore switch to Markov Chain Monte Carlo (MCMC) methods to obtain samples from each agent's posteriors, and approximate the expectations in Eq.~\ref{eq:marginalized} by summing over these samples. 
Because we are emphasizing a set of phenomena where our model makes qualitatively different predictions than previous models, our goal in this paper is to illustrate and evaluate these \emph{qualitative} predictions rather than provide exact quantitative fits to empirical data.
As such, we proceed by examining predictions for a regime of parameter values ($w_S, w_L, w_C,\beta$) that help distinguish our predictions from other accounts.

\section{Appendix B: Alternative lexical representations}

In this section, we re-consider two specific choices we made about how to represent lexical meanings.

First, for simplicity and consistency with earlier models of Bayesian word learning, we adopted a traditional truth-conditional representation of lexical meaning throughout the paper. 
Each word in the lexicon is mapped to a single `concept' to , e.g. $w_1 = `blue square'$, where this utterance is true of objects the fall in the given concept, and false otherwise. 
The inference problem over lexicons therefore requires searching over this discrete space of word-concept mappings. 
However, it is important to emphasize that our model is entirely consistent with alternative lexical representations.

For example, for some settings, a \emph{continuous, real-valued} representation may be preferred, or a higher-dimensional vector representation.
Rather than assigning each word a discrete concept in the lexicon, we may simply assign each word-object pair $(w_i,o_j)$ a scalar meaning representing the extent to which word $w_i$ applies to object $o_j$, such that $\phi$ is a real-valued matrix:
$$
\phi = \begin{bmatrix}
\phi^{(11)} & \phi^{(12)} & \cdots & \phi^{(1i)} \\
\phi^{(21)} & \phi^{(22)} & \cdots & \phi^{(2j)} \\
\vdots & \vdots &\ddots & \vdots \\
\phi^{(j1)} & \phi^{(j2)} & \cdots & \phi^{(ij)} 
\end{bmatrix}
$$
and $\mathcal{L}_\phi(w_i, o_j) = \phi^{(ij)}$. 
In this case, rather than discrete categorical priors over meanings, we may place Gaussian priors over the entries of this matrix:
\begin{align}
\Theta^{(ij)} & \sim \mathcal{N}(0,1)\nonumber \\
\phi^{(ij)} & \sim \mathcal{N}(\Theta^{(ij)} | 1)\nonumber
\end{align}
We have previously achieved similar results using this alternative lexical representations in earlier iteration of this manuscript \cite{hawkins_convention-formation_2017,hawkins2020generalizing}, although deriving predictions required variational inference techniques rather than Markov Chain Monte Carlo. 
Such optimization-based inference techniques may also provide the most promising path for extending our adaptive model to larger language models, including neural networks that operate over continuous spaces of image pixels and natural language embeddings \cite{hawkins2019continual}.

\section*{Appendix C: Alternative lexical priors}

A variety of priors have been proposed for probabilistic models of language learning and convention formation, which build in stronger or weaker assumptions about the structure of the lexicon.
We used a simplicity prior over lexicons that strictly partition the set of referents, ensuring that every object must be in the extension of exactly one word \cite<e.g.>{carr2020simplicity}. 
To explore the robustness of our results in \textbf{P3} to alternative choices of lexical priors, we also considered a weaker prior where the space of possible lexicons allows any denotation to be assigned to any word, including highly redundant and overlapping lexicons (where every object is in the extension of every word), and highly degenerate lexicons (where every word has a completely empty extension).
Given this more unconstrained space of lexicons, the simplest way to penalize complexity is to define the size of the lexicon $|\phi|$ as the \emph{total extension size}, i.e. the summed extensions of all terms.
In this case, favoring simpler word meanings also necessarily favors smaller lexicons. 

\begin{figure*}
\centering
    \includegraphics[scale=.7]{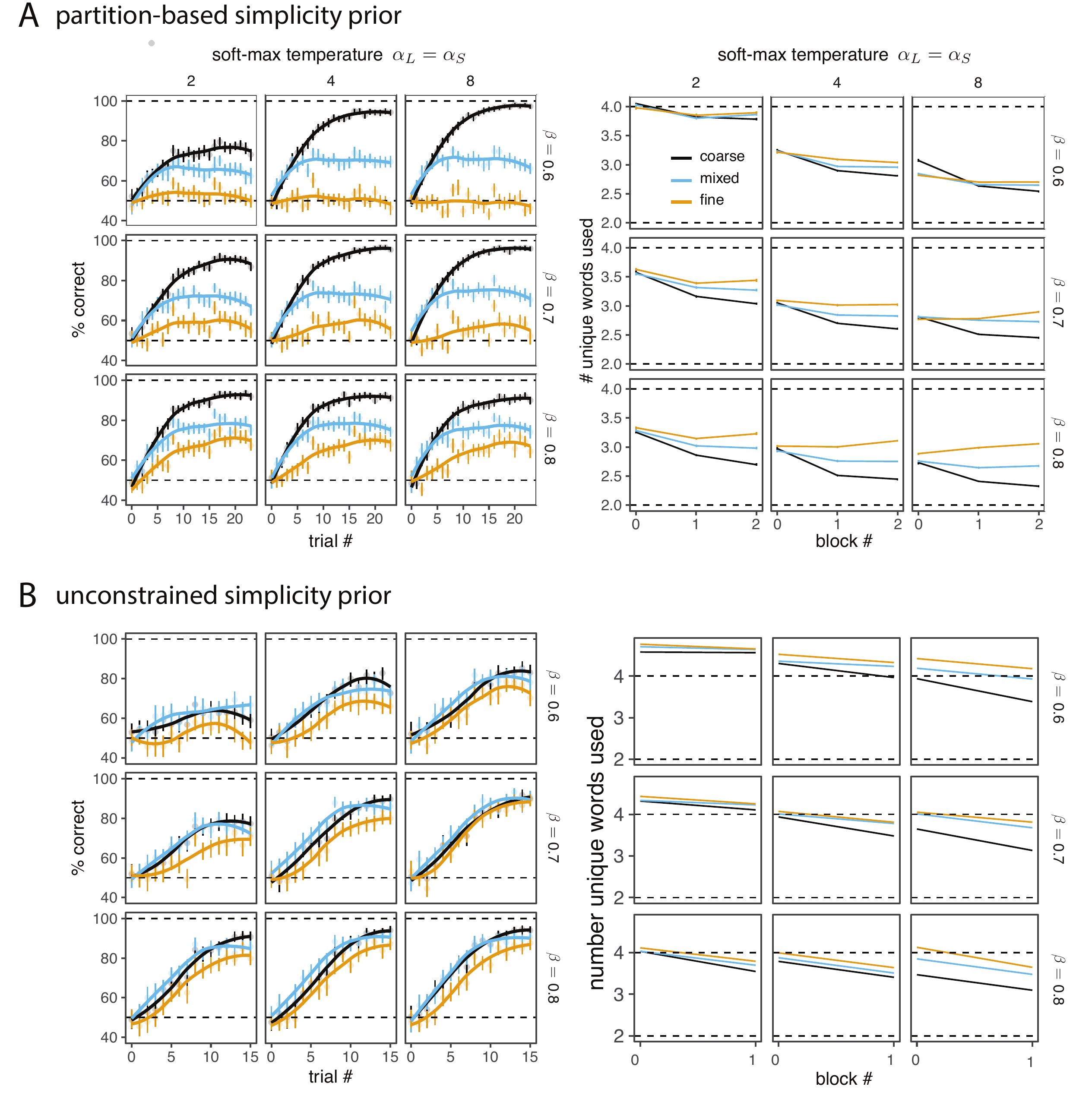}
  \caption{Simulation results for \textbf{P3} using different values of $\alpha$ and $\beta$ for (A) partition-based simplicity prior, and (B) alternative unconstrained simplicity prior. Simulations were run for fewer trials in (B). Overall, we observe similar qualitative predictions for the difference between the coarse and fine condition, although the mixture condition is more sensitive to parameters and priors.}
  \label{fig:partitionPrior}
\end{figure*}

To illustrate this property of the weaker prior, consider a reference game with a fixed set of two referents: a blue square and a red square. 
Then the meaning of a given utterance $u_i$ can either have extension size two (applying both objects, effectively meaning ‘square’), size one (applying only to ‘blue square’ or only to ‘red square’), or size zero (applying to neither of the objects). 
Now suppose there are 16 possible utterances. 
Then the lexicon with highest prior probability is the one where every utterance has an empty extension ($|\phi| = 16 \cdot 0 = 0$), which is also the smallest lexicon (an effective vocabulary size of 0). 
The lexicon with lowest prior probability is the one where every utterance has the maximal extension ($|\phi| = 16 \cdot 2 = 32$), which is also the largest lexicon (an effective vocabulary size of 16). 
Removing a single word from this maximal lexicon would reduce the size of the lexicon (e.g. 15 words instead of 16) and also reduce the total size of the words' extensions (e.g. taking a word with the maximal extension of size two and replacing it with the minimal extension of size zero: $|\phi| = 15 \cdot 2 + 1 \cdot 0 = 30$), which would slightly increase its prior probability. 
Of course, this scheme also makes it possible to have a lexicon with smaller word meanings but not a smaller lexicon (e.g. for a lexicon where all 16 words have an extension size of 1, we have $|\phi| = 16 \cdot 1 = 16$, but the effective vocabulary size is still 16). 
So this prior straightforwardly encodes a preference for lexicons with fewer words, but gives partial credit when the words have ‘simpler’ meanings, breaking ties between lexicons with the same number of words.
Simulation results using this prior are shown in Fig.~\ref{fig:partitionPrior}B. 

\begin{figure}[h]
\centering
    \includegraphics[scale=.5]{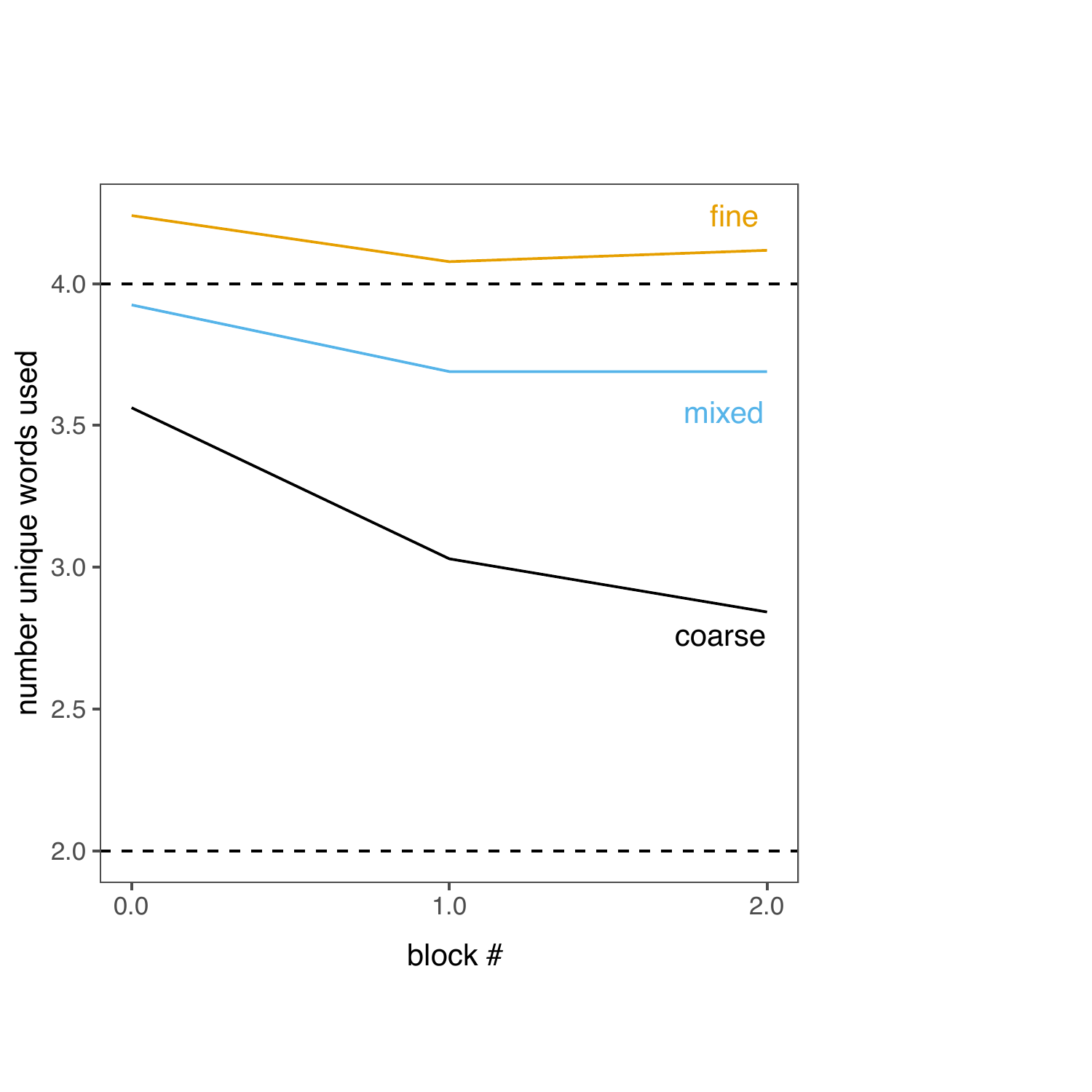}
  \caption{Simulation results for \textbf{P3} using a full-coverage lexical prior that disallows degenerate lexicons ($\alpha_S = \alpha_L = 6, \beta = 0.6$)}
  \label{fig:partitionMiddle}
\end{figure}

Note that other choices are possible, such as those only enforcing mutual exclusivity, or only the principle of contrast \cite{clark1987principle}.
In the context of \textbf{P3}, we expect these choices to primarily affect choices on early trials.
Some priors may predict an \emph{agglomerative} form of learning where all conditions begin using fine-grained language and then the \emph{coarse} condition gradually collapses down to a more minimal lexicon, while others predict a \emph{divisive} form where all conditions begin using coarse-grained language and then the \emph{fine} condition gradually introduces words with more refined meanings. 
For example, we one qualitative feature of our empirical results in \textbf{P3} (see Fig.~\ref{fig:sec2Results}D) is that participants apparently begin using more unique terms at the outset, and remain constant at that large number of unique terms in the \emph{fine} and \emph{mixed} conditions but gradually whittle away their vocabulary size in the \emph{coarse} condition.
Because we noticed that neither of the priors considered in Fig.~\ref{fig:partitionPrior} displayed this pattern, we considered a third possible prior. 
This prior enforced full coverage over all meanings (i.e. disallowed `degenerate' lexicons where some objects are not in the extension of any words at all), unlike the unconstrained prior, but otherwise allowed redundancy (i.e. some objects were in the extension of multiple words), unlike the partition-based prior.
This prior gave rise to a qualitatively more similar pattern of lexical convergence (see Fig.~\ref{fig:partitionMiddle}).

 \begin{figure*}
\centering
    \includegraphics[scale=1.2]{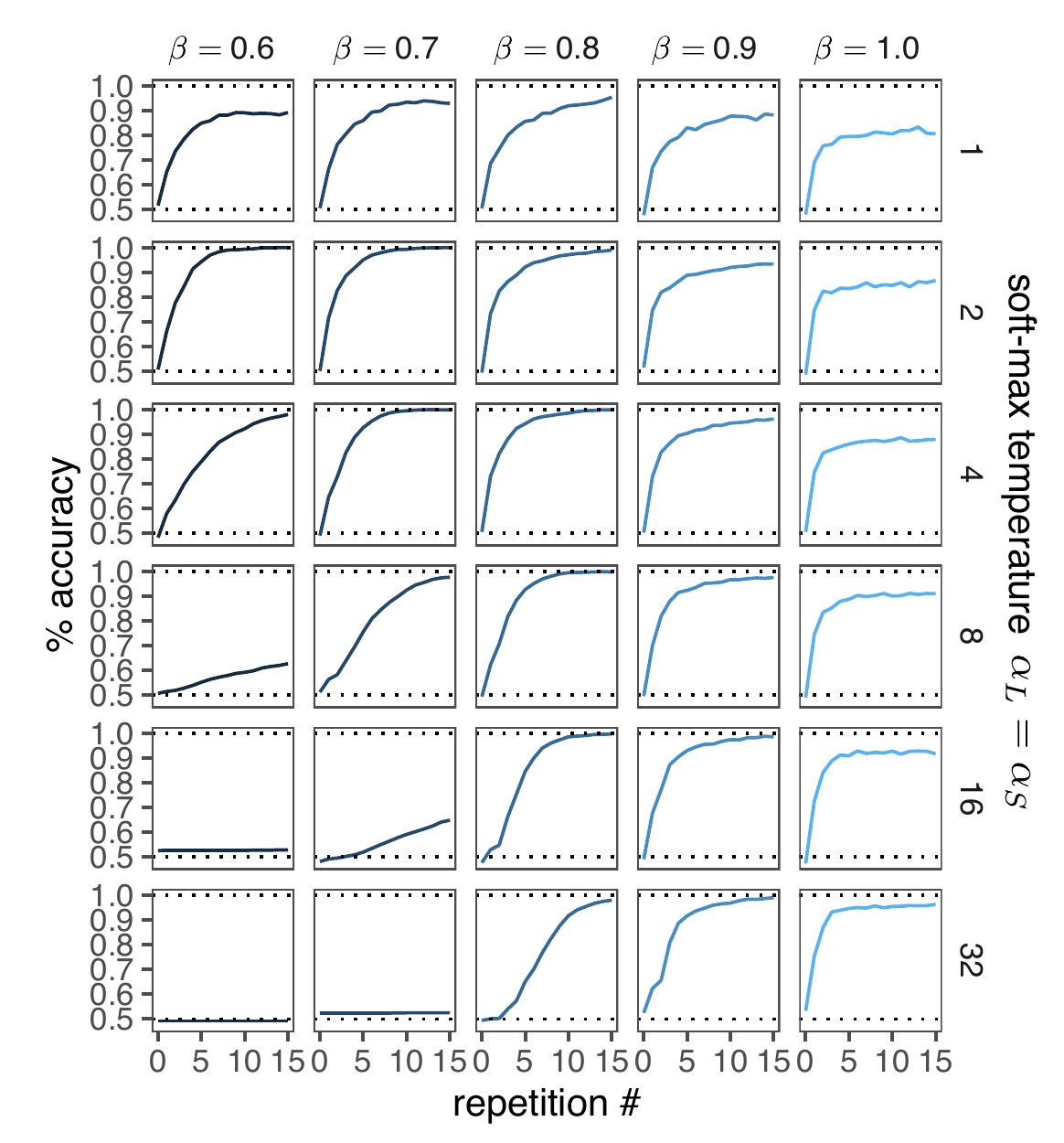}
  \caption{Coordination success (simulation 1.1) across a range of parameter values. Columns represent memory discount parameter $\beta$, and rows represent the agents' soft-max optimality parameters, where we set $\alpha_S=\alpha_L$. Communicative success is achieved under a wide range of settings, but convergence is limited in some regimes. For example, at high values of $\beta$, with no ability to discount prior evidence, accuracy rises quickly but asymptotes below perfect coordination; at low $\alpha$, inferences are slightly weaker and agent actions are noisier, slowing convergence; finally, at low values of $\beta$, when prior evidence is forgotten too quickly, convergence interacts with $\alpha$: the latest evidence may overwhelm all prior evidence, preventing the accumulation of shared history. The agent noise model is set to $\epsilon = 0.01$ in all simulations.}
  \label{fig:arbitrariness_grid}
\end{figure*}

 \begin{figure*}
\centering
    \includegraphics[scale=1.2]{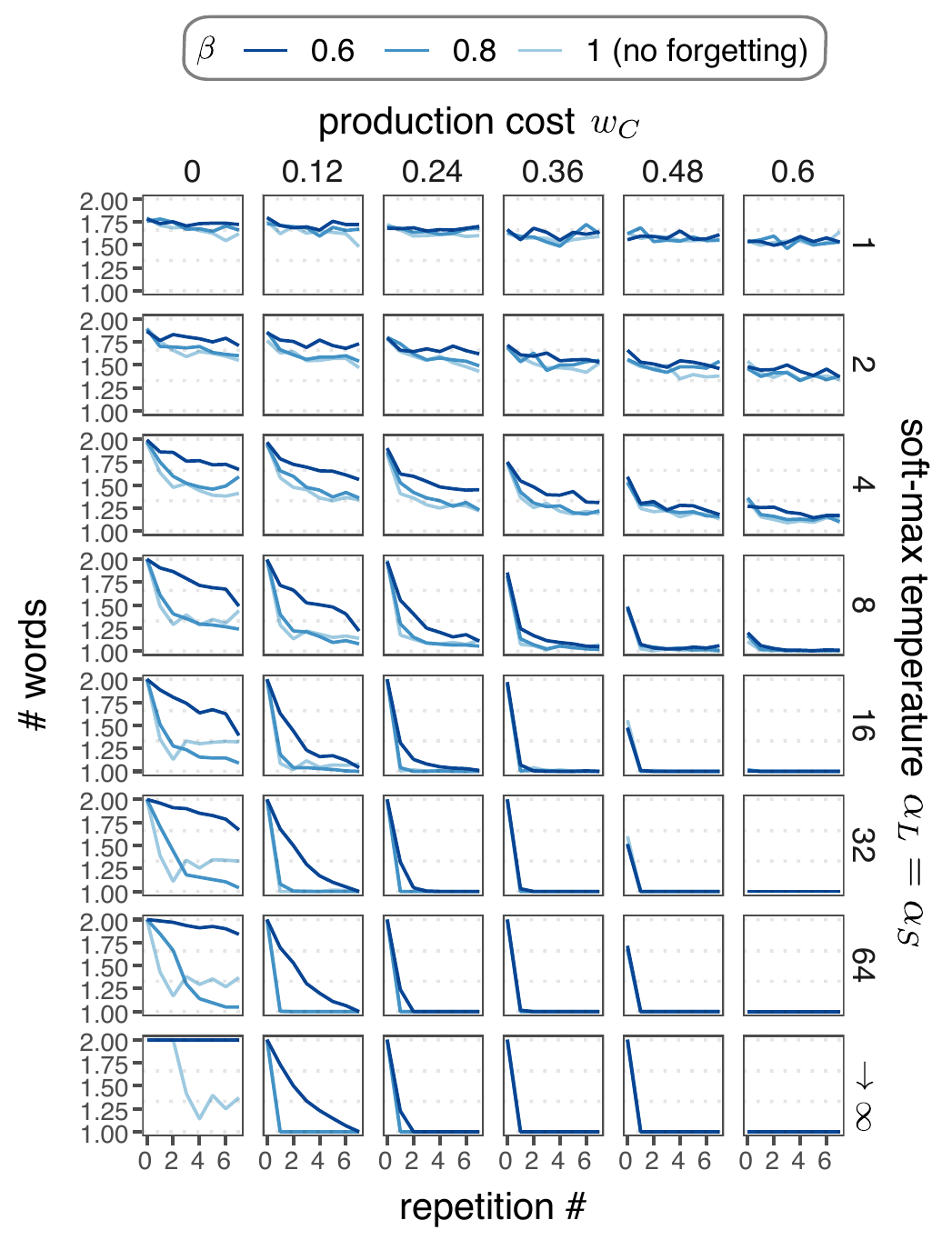}
  \caption{Speaker efficiency (simulation 1.2) across a range of parameter values representing different weights on informativity and cost. Rows represent agent soft-max optimality $\alpha_S = \alpha_L$, columns represent costs $w_C$, and different memory discount factors $\beta$ shown in different colors. Agents converge on more efficient ad hoc conventions for a wide regime of parameters. When utterance production cost $w_C$ is more heavily weighted relative to informativity, the speaker is less likely to produce longer utterances, even at the beginning of the interaction; when optimality $\alpha_S, \alpha_L$ is higher, and the speaker maximizes utility, we observe faster reduction and more categorical behavior. Note that as $\alpha\rightarrow\infty$, utterances only become shorter at $w_C=0$ in the absence of forgetting. In this case, the shorter utterances approach the exact same utility as the longer utterance, and the speaker reaches equilibrium simply sampling among them at random (i.e. choosing the longer utterance with 1/3 probability and each of the shorter utterances with 1/3 probability).}
  \label{fig:conjunction_grid}
\end{figure*}

 \begin{figure*}
\centering
    \includegraphics[scale=.8]{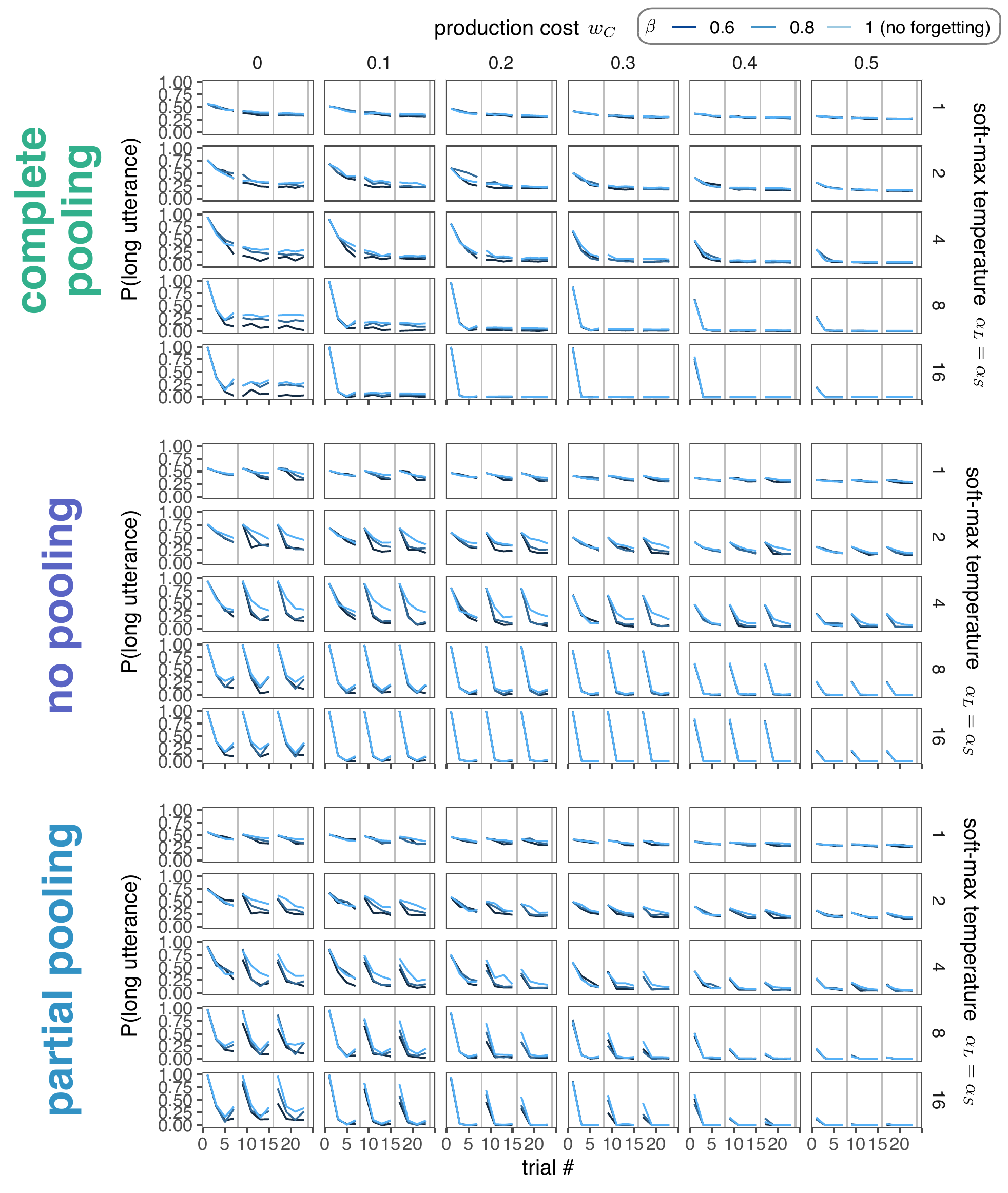}
  \caption{Speaker efficiency simulations for \textbf{P2} across a larger parameter regime. We examine the behavior of complete-pooling, no-pooling, and partial-pooling models, where rows represent agent soft-max optimality $\alpha_S = \alpha_L$, columns represent cost weight $w_c$, and colors represent memory discount parameter $\beta$.}
  \label{fig:partnerspecificity_grid}
\end{figure*}

\begin{figure*}
\includegraphics[scale=1.2]{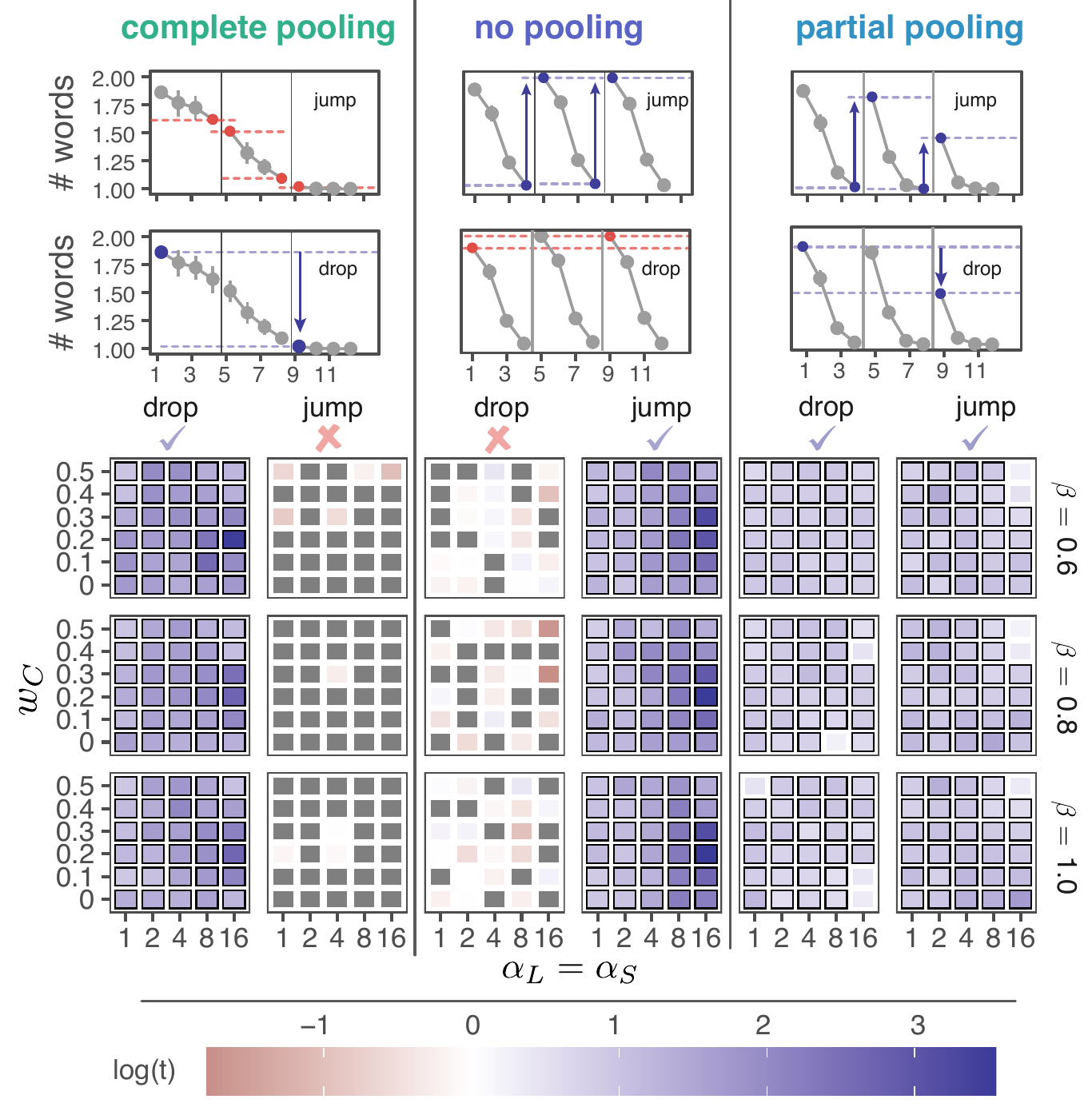}
\caption{Qualitative predictions of our three models for \textbf{P2}. Across a wide range of parameter values, only the hierarchical model consistently produces both qualitative phenomena of interest: reversion to the prior at partner boundaries (i.e. a ``jump'') and gradual generalization across partners (i.e. a ``drop''). Approximately $N=10$ simulations used to compute $t$-statistic in each cell. Cells marked with black boxes are significantly different from a null effect of 0 change, $p<0.005$.}
\label{fig:generalization_modelcomparison}
\end{figure*}

 \begin{figure*}[h!]
\centering
    \includegraphics[scale=.9]{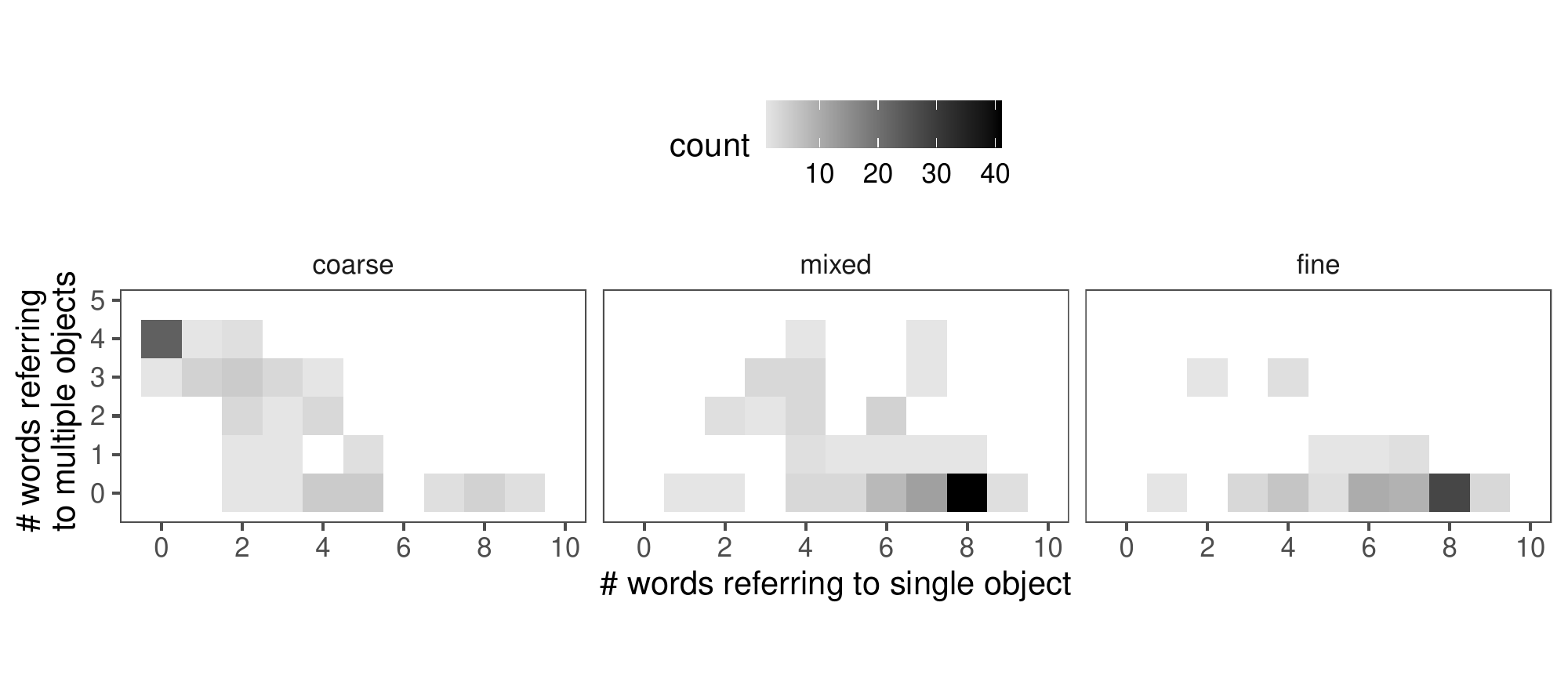}
  \caption{Empirical mixtures of terms reported by participants in \textbf{P3}. While the modal lexicon in the coarse condition contained 0 specific terms and 4 more general terms (32\% of participants) and the modal lexicon in the mixture and fine conditions contained 8 specific terms and 0 more general terms (42\% and 38\% of participants, respectively), many participants reported a mixture of abstract and specific terms.}
  \label{fig:mixtureOfTerms}
\end{figure*}

\end{document}